\documentclass{article}
\PassOptionsToPackage{table}{xcolor}
\usepackage{iclr2027_conference,times}

\usepackage{hyperref}
\hypersetup{hidelinks}

\AtBeginDocument{\fancyfoot[C]{\number\value{page}}}
\usepackage{url}
\usepackage[T1]{fontenc}
\usepackage{booktabs,graphicx,xcolor}
\usepackage{placeins}
\usepackage{float}

\usepackage{amsmath,amssymb,amsfonts}
\usepackage{tabularx,array}
\usepackage{xspace}

\usepackage[capitalize]{cleveref}
\crefname{section}{Sec.}{Secs.}
\Crefname{section}{Section}{Sections}
\crefname{table}{Tab.}{Tabs.}
\Crefname{table}{Table}{Tables}

\newcommand{\bench}{AutoOntologyBench\xspace}

\newcommand{\IllustrativeCaption}{}

\newcommand{\mSecEval}{Sec.~4 of the main paper}
\newcommand{\mSecRuns}{Sec.~4.1 of the main paper}
\newcommand{\mSecCond}{Sec.~4.2 of the main paper}
\newcommand{\mEqAdmit}{Eq.~(1) of the main paper}
\newcommand{\mEqPaired}{Eq.~(6) of the main paper}
\newcommand{\mFigHoney}{Fig.~2 of the main paper}
\newcommand{\mFigResults}{Fig.~3 of the main paper}
\newcommand{\mTabMain}{Table~1 of the main paper}
\newcommand{\mTabRep}{Table~2 of the main paper}

\title{AutoOntologyBench:\\A Task-Grounded Benchmark for Agentic Knowledge Engineering}

\author{
\small
\begin{tabular}{@{}l@{}}
\textbf{Minye Shao\textsuperscript{1,4}},
\textbf{Chenghao Xiao\textsuperscript{2}},
\textbf{Xingyue Zhao\textsuperscript{3,4}},
\textbf{Yang Long\textsuperscript{1}}
\end{tabular}
\\[10pt]
{\small
\begin{tabular}{@{}l@{}}
\textsuperscript{1}Durham University\\
\textsuperscript{2}Shanghai University of Finance and Economics\\
\textsuperscript{3}Peking Union Medical College Hospital, Peking Union Medical College\\
\textsuperscript{4}Alibaba DAMO Academy, Alibaba Group
\end{tabular}}
}

\iclrfinalcopy
\begin{document}
\maketitle

\begin{abstract}
Foundation-model agents are increasingly extending beyond software engineering (SWE) into \textbf{autonomous scientific research}, where acquiring and integrating domain knowledge remains a critical challenge. Recent industrial advances have increasingly adopted knowledge graphs and ontology-based semantic layers to organize heterogeneous knowledge into reusable representations. We ask: \textbf{Can autonomous agents construct ontologies that help them solve real tasks, and do they offer measurable benefits over alternative knowledge representations?} Existing benchmarks evaluate ontology correctness or agent task success in isolation, leaving the value of autonomously acquired knowledge unresolved. We therefore introduce \bench, a task-grounded benchmark comprising \textbf{19,947} audited tasks grounded in \textbf{18 real-world} knowledge sources across \textbf{7 domains}, with source-native verification.
Spanning Application, Repair, Organization, and Construction,
\textbf{\bench{}} compares nine frontier-model agents under matched knowledge representations.
Layered on retrieval, agent-built ontologies add 6.1 points of task success, yet trail an expert-derived reference by 13.0,
identifying autonomous knowledge acquisition as the bottleneck to scalable scientific research.
Code and resources are available at \url{https://github.com/VinyehShaw/AutoOntologyBench}.
\end{abstract}

\section{Introduction}
\label{sec:intro}

Foundation-model agents have made rapid progress in software engineering, where access to source code, executable tools, and automated tests enables iterative problem solving and verification~\cite{swebench,react}. These advances have inspired \emph{AutoResearch}, extending autonomous experimentation beyond software engineering toward scientific discovery and other knowledge-intensive domains~\cite{karpathy2026autoresearch}. Comparable reliability has not followed: recent benchmarks report large gaps in scientific reasoning, experimental exploration, and the execution of domain-specific research workflows~\cite{chen2025scienceagentbench,discoveryworld,liu2026sciagentarena}. A central difficulty lies in the nature of domain knowledge. Unlike software repositories, where task-relevant information is often directly inspectable and executable, scientific and industrial knowledge is distributed across heterogeneous databases, literature, technical specifications, and independently maintained rules. Solving a task may require reconciling inconsistent identifiers, recovering implicit constraints, or integrating relationships across multiple sources. Consequently, extending autonomous research beyond software engineering requires agents not only to reason and act, but also to \emph{acquire, organize, and reuse domain knowledge} that is not explicitly available in their immediate context.

Knowledge graphs and ontologies are a long-standing way to organize such knowledge. Knowledge graphs represent entities and their relationships, while ontologies make domain concepts, relations, and constraints explicit, providing a semantic structure for consistent interpretation and reuse~\cite{hogan2021knowledgegraphs,gruber1993,studer1998}. Recently, NVIDIA Auto Ontology has illustrated growing industrial interest in automatically constructing ontology-based semantic layers from enterprise metadata, enabling AI agents to access structured domain knowledge~\cite{nvidia-ao}. Related developments from Palantir and Amazon Web Services (AWS) further reflect the adoption of ontology-driven knowledge integration and agent interaction~\cite{palantir,aws-coa}. However, the practical adoption of ontology-based systems does not establish that general-purpose agents can construct them reliably. More importantly, ontologies are only one way to support knowledge-intensive agents. Alternative approaches include direct reasoning and tool interaction~\cite{react}, retrieval-augmented generation (RAG)~\cite{rag}, and writing knowledge directly as code or domain-specific rules~\cite{codeact}. Each involves different trade-offs in knowledge acquisition, semantic expressiveness, and reuse, and they are not mutually exclusive: an ontology can organize what retrieval finds rather than replace it. This raises the question: \emph{Can autonomous agents construct useful ontologies from incomplete evidence, and when does doing so provide measurable advantages over alternative knowledge representations?} Existing ontology evaluations primarily measure artifact correctness, consistency, and semantic alignment~\cite{cq4oe,oaei2025}, whereas agent benchmarks emphasize downstream task completion~\cite{swebench,taubench,autodatabench2026}. Neither connects the quality of autonomously acquired knowledge to its contribution to verified task outcomes, so the value of agent-generated ontologies remains unmeasured.

\begin{figure}[t]
    \centering
    \includegraphics[width=\linewidth]{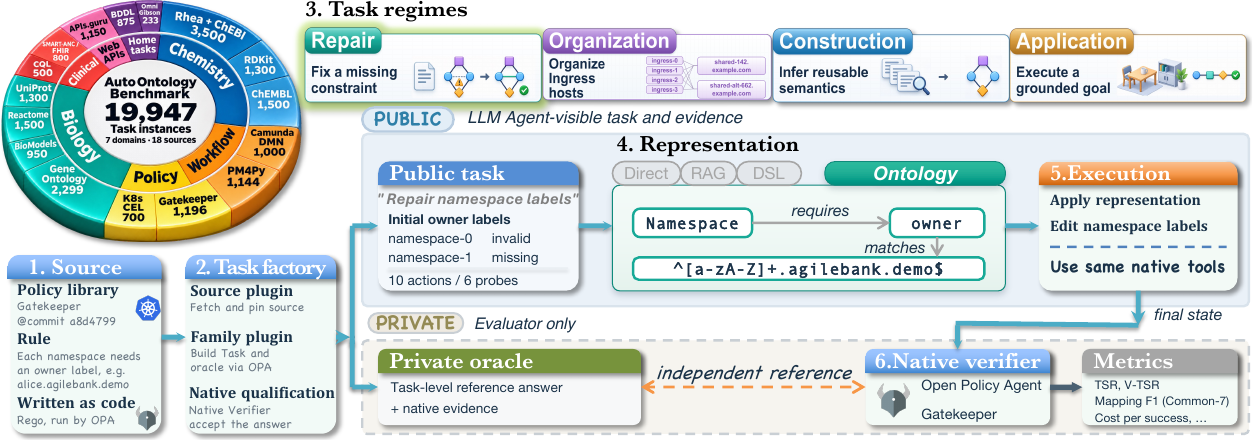}
    \caption{\textbf{Overview of \bench, illustrated with one Gatekeeper task.} Left: the audited bank, 19{,}947 tasks in 16 lanes, with seven domains (inner ring) and pinned sources (outer ring). Right: (1)~a pinned source, here a Kubernetes policy rule that every namespace needs an owner label; (2)~the task factory fetches and pins the source, builds a task and its private oracle by solving it natively with the Open Policy Agent (OPA)~\cite{opa}, and qualifies it through native verification; (3)~the task's regime, here Repair, and the public task the agent sees; (4)~the knowledge representation the agent builds, here an ontology, with Direct, RAG, and a native DSL as the alternatives; (5)~execution with the same native tools in every condition; (6)~the native verifier checks the final state against the private oracle, and we report task success and valid task success.}
    \label{fig:framework}
\end{figure}

We introduce \bench, a task-grounded benchmark for evaluating \emph{agentic knowledge engineering through downstream utility}, as illustrated in \cref{fig:framework}, which follows one task from a pinned source (1) through the task factory (2), its regime (3), the knowledge representation (4), and execution (5) to the native verifier (6). Grounded in 18 real-world, versioned knowledge sources spanning seven domains, \bench comprises 19{,}947 audited task instances derived from 4{,}511 base goals and 325 task families across 16 native-verifiable task environments. The benchmark encompasses knowledge application, ontology repair, cross-source organization, and ontology construction, with task outcomes independently evaluated using source-specific native oracles. We evaluate nine frontier models under matched knowledge representations, measuring native-verified task success, mapping quality together with its scoring coverage, cost per valid success, and the effect of controlled semantic interventions. \bench lets us test \emph{whether, when, and why} autonomously constructed knowledge improves downstream agent performance, including whether an agent-built ontology adds value on top of retrieval. Our contributions are threefold:

\begin{itemize}
    \item \textbf{A task-grounded benchmark for agent-built ontologies.}
    To our knowledge, \bench{} is the first benchmark to evaluate agent-built ontologies through independently verified downstream task success, moving beyond artifact correctness to measure their actual utility for autonomous agents.

    \item \textbf{Establishing a source-grounded benchmark for agentic knowledge engineering.}
    We construct \bench{} from real-world, versioned domain knowledge, comprising 19{,}947 audited tasks across 16 verifiable environments. Native oracles, each validated by an admission audit before release, provide reliable evaluation across knowledge application, repair, organization, and construction.

    \item \textbf{Revisiting the value of structured knowledge for autonomous agents.}
    We compare agent-built ontologies with direct tool use, retrieval, and lane-native DSL artifacts, both alone and layered on retrieval, across nine frontier models and seven domains. Semantic interventions and per-domain analysis show when structured knowledge improves task performance and when alternative representations remain preferable.
\end{itemize}

\section{Related Work}
\label{sec:related}
\paragraph{Agent benchmarks with verifiable outcomes.}
SWE-bench grades repository patches with executable tests~\cite{swebench}, AgentBench and WebArena evaluate agents in interactive environments~\cite{agentbench,webarena}, $\tau$-bench checks the final database state of a tool--user dialogue~\cite{taubench}, and DiscoveryWorld separates task completion from explanatory knowledge~\cite{discoveryworld}. AutoDataBench, our closest relative, fixes all non-data factors to measure how agents diagnose, organize, and construct training data~\cite{autodatabench2026}. \bench shares this outcome-first, regime-structured design, but the artifact under test is an ontology the agent builds and then acts through, and ground truth comes from each source's own executor.

\paragraph{Knowledge representations for agents.}
Agents can keep knowledge in the context window through tool use~\cite{react,toolformer}, fetch it at query time~\cite{rag}, or write it down as code~\cite{pal,codeact,voyager}. Our Native DSL baseline keeps the last idea but restricts it to each source's declarative language, so the artifact cannot read hidden state. Graph-based retrieval builds entity graphs over a corpus~\cite{edge2024graphrag,sarmah2024hybridrag,peng2024graphragsurvey}, and LLM-based ontology learning extracts terms, taxonomies, and schema-conformant knowledge bases~\cite{babaei2023llms4ol,caufield2024spires}. These works score answers against text or score the extracted structure itself. Our RAG + Ontology condition shares GraphRAG's premise that structure helps retrieval, but judges the ontology only by whether a native verifier accepts the downstream answer.

\paragraph{Ontologies and knowledge graphs for agents.}
Ontologies make concepts, relations, and constraints explicit and shareable~\cite{gruber1993,studer1998}, and knowledge graphs ground language-model reasoning~\cite{pan2024unifying,tog}. EvoOntology evolves an ontology layer for data agents~\cite{evoontology}, OaK refines task-oriented ontologies during execution~\cite{oak}, NVIDIA's Auto Ontology Eval compares schema-only with ontology-grounded Text-to-SQL~\cite{nvidia-ontology-eval}, and commercial platforms deploy ontology agents~\cite{fabric-ontology-agent,palantir-ontology-overview}. None of these isolates a privileged verifier, compares against lane-native and retrieval baselines on the same tasks, or tests whether a specific semantic fault causes a specific failure. Artifact measures such as OWL/SHACL validity and OAEI alignment~\cite{oaei2025} serve here only as diagnostics.

\section{AutoOntologyBench}
\label{sec:design}

\bench turns real, versioned domain knowledge into tasks that each source's own engine can grade. \Cref{fig:framework} follows one Gatekeeper task, repairing the owner labels of two Kubernetes namespaces, through six steps; this section covers steps 1 to 3 (where tasks come from and what they look like) and \cref{sec:evaluation} steps 4 to 6 (how an agent is run and scored).

\noindent\textbf{Sources and lanes.}
Every task starts from a \emph{source snapshot}, one upstream revision pinned by commit or release (step~1 of \cref{fig:framework}). In the figure the snapshot is the Gatekeeper policy library~\cite{gatekeeper}, whose rule that every namespace needs an owner label is written in Rego, the policy language of the Open Policy Agent (OPA)~\cite{opa}, and executed by OPA. \bench uses 18 snapshots in seven domains: \emph{Policy} (Gatekeeper, Kubernetes Common Expression Language (CEL)~\cite{k8scel}), \emph{Workflow} (Camunda Decision Model and Notation (DMN)~\cite{camunda}, PM4Py~\cite{pm4py}), \emph{Chemistry} (Rhea, ChEBI~\cite{rhea,chebi}, RDKit~\cite{rdkit}, ChEMBL~\cite{chembl}), \emph{Biology} (Gene Ontology~\cite{go}, BioModels~\cite{biomodels,sbml}, Reactome~\cite{reactome}, UniProt~\cite{uniprot}), \emph{Clinical} (Clinical Quality Language (CQL)~\cite{cql}, World Health Organization (WHO) SMART antenatal care guidelines (SMART-ANC)~\cite{smartanc}), \emph{Home tasks} (BEHAVIOR-1K, OmniGibson~\cite{bddl,behavior1k}), and \emph{Web APIs} (APIs.guru~\cite{apisguru}). A \emph{lane} connects one or two snapshots, for example Rhea with ChEBI, to a public task interface and a native verifier. This gives 16 lanes, numbered L01 to L16 in the supplementary material.

\noindent\textbf{Task factory.}
The factory (step~2) runs three stages per source. A \emph{source plugin} fetches and pins the snapshot. A \emph{family plugin} turns a \emph{task family}, a reusable template with its verifier branch, into concrete tasks: it chooses entities and an initial state, computes a reference answer with the native tools and the private source rules, and splits the result into a \emph{public task} (goal, initial state, public evidence, permitted tools, and budgets) and a private oracle $O_{\mathcal T}$ (the reference answer, rules, and hidden tests). The agent sees only the public task, which starts unsolved; the answer exists only inside the factory. \emph{Native qualification} then keeps a task only if the lane's native verifier accepts the reference answer and the oracle can be recomputed from the pinned snapshot (see below). The native verifier $V_{\mathcal T}(z;O_{\mathcal T})\in\{\mathrm{PASS},\mathrm{FAIL}\}$ checks a submitted answer or final state $z$ against $O_{\mathcal T}$ by running the lane's native executor $E_w$ on the pinned snapshots $w$. Tasks are counted at three levels. A \emph{base goal} is one objective inside a family, and a \emph{task instance} binds a base goal to concrete entities, initial state, and parameters; instances of one base goal share the goal text but differ in entities, initial values, and oracle. The 19{,}947 instances therefore cover 4{,}511 distinct objectives in 325 families. For example, the Gatekeeper label-repair family has 56 instances spanning 15 base goals, and the six instances of one of them differ in namespace names and initial labels. An LLM phrases each goal in natural language, and any new goal structure it proposes is re-instantiated and checked against its source. Deduplication, solvability, nontriviality, and leakage checks run before the final split, which groups tasks by source mechanism rather than wording. Where a lane supports it, budgeted probes let the agent query the native executor.

\noindent\textbf{Oracle qualification.}
Each candidate is checked by up to five gates, and only PASS admits it; a gate that has not run counts as not passed. Writing $\mathcal Q$ for the gates that apply to the candidate's lane,
\begin{equation}
\begin{aligned}
\mathrm{Admit}(\mathcal T)&=\prod_{q\in\mathcal Q}\mathbb 1\bigl[q(\mathcal T)=\mathrm{PASS}\bigr],\\
\mathcal Q&\subseteq\{\mathrm{src},\mathrm{det},\mathrm{wit},\mathrm{mut},\mathrm{goal}\}.
\end{aligned}
\label{eq:admit}
\end{equation}
Here $\mathrm{src}$ checks the pinned source and adapter; $\mathrm{det}$ checks that repeated native executions agree; $\mathrm{wit}$ requires a valid reference answer to be accepted and an incorrect witness to be rejected; $\mathrm{mut}$ requires source-changing mutations to change the verdict as expected; and $\mathrm{goal}$ checks that the task objective matches the native outcome. Of 20{,}000 candidates, 19{,}947 are admitted and 53 excluded (35 goal mismatches, 7 mutation-insensitive cases, 5 missing positive witnesses, 4 duplicates, and 2 engine errors). A second, independent engine re-verifies 19{,}210 of the admitted tasks; for the other 737 it is not available (NOT\_AVAILABLE, neither PASS nor FAIL). Most lanes are verified by an external engine such as OPA, an OWL reasoner, or a simulator, and the knowledge-base lanes by exact queries over the pinned release (see the supplementary material).

\begin{figure}[t]
\centering
\includegraphics[width=0.55\linewidth]{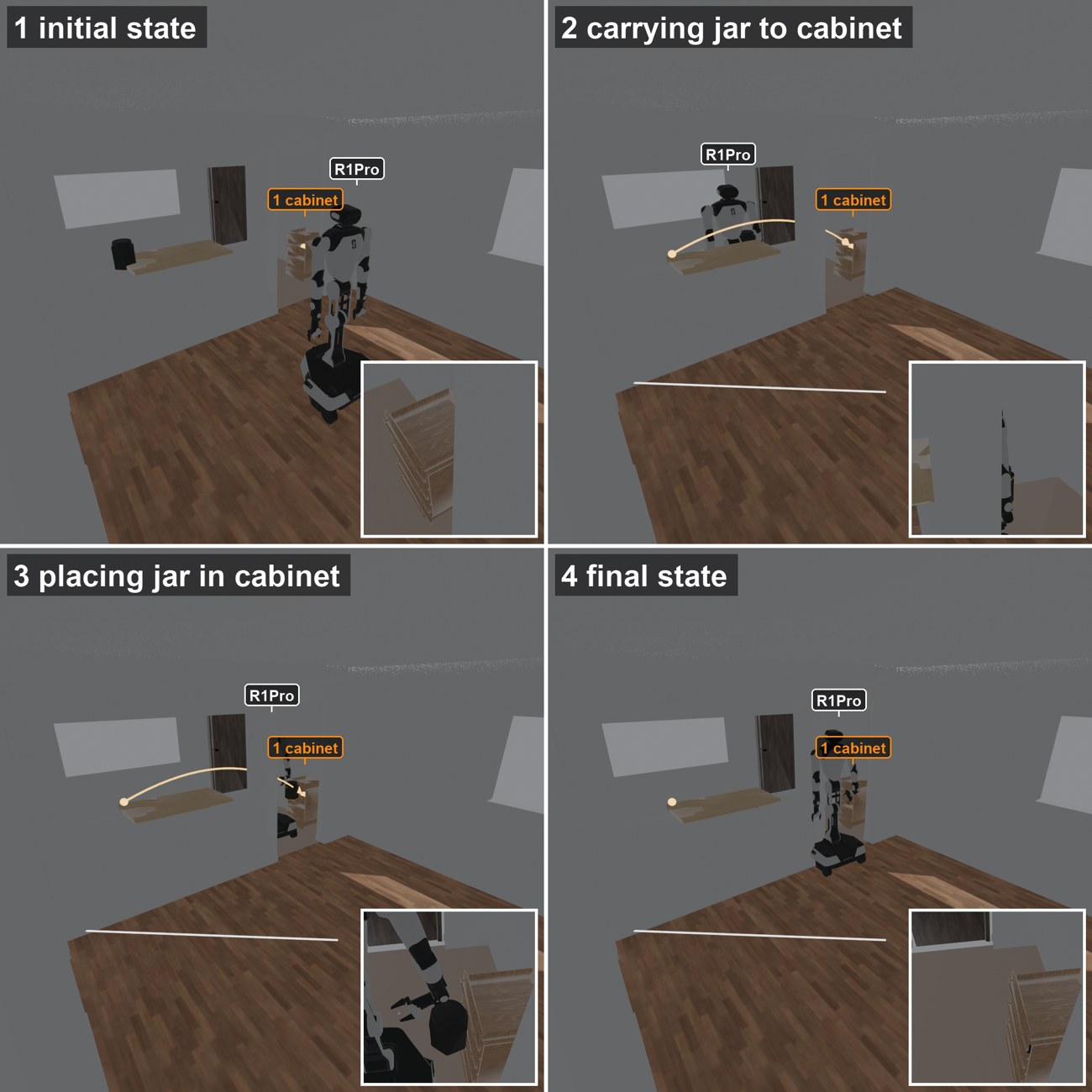}
\caption{\textbf{An example of the Application regime: an embodied task from the Home-task lane (L11).} A mobile manipulator in a BEHAVIOR-1K kitchen receives ``Put the jar of honey inside the cabinet'' and submits a program of symbolic primitives. The oracle replays the program in OmniGibson and checks the goal, written in the BEHAVIOR Domain Definition Language (BDDL), at the final state. Placement is replayed symbolically, so the frames are re-staged for readability and do not show contact-level physical feasibility.}
\label{fig:honey}
\end{figure}

\noindent\textbf{Task regimes.}
Each task carries one of four regime labels (step~3), assigned at construction. \emph{Application} (7{,}033 tasks) starts from an ordinary input and applies existing source knowledge to answer a query or execute a goal, as in the embodied task of \cref{fig:honey}. \emph{Repair} (6{,}100) starts from a defective artifact and returns a repaired one that leaves unaffected knowledge intact; the running example, repairing two owner labels, is a Repair task, so its card is highlighted in \cref{fig:framework}. \emph{Organization} (5{,}319) starts from empty mapping stubs and aligns entities and relationships across independent sources, such as the missing Rhea--ChEBI cross-references. \emph{Construction} (1{,}495) starts from partial evidence and builds a structured output that satisfies source-defined constraints, including graphs, simulation parameters, and API instances; its output need not be a complete ontology. One worked example per lane is in the supplementary material.

\noindent\textbf{Task difficulty.}
Each task is rated on four source-dependent factors that describe the structure of its goal rather than its wording: semantic dependency, planning horizon, evidence ambiguity, and cross-source composition. The tasks comprise 4{,}767 low-, 8{,}446 medium-, and 6{,}734 high-complexity instances (see the supplementary material). \textbf{Extensibility.}
New sources can enter the benchmark through version pinning, native executor integration, task-family specification, and oracle qualification.

\section{Evaluation}
\label{sec:evaluation}
This section covers steps 4 to 6 of \cref{fig:framework}: the representation in which an agent stores what it learns (\cref{sec:conditions}), the execution and verification of a run (\cref{sec:success}), and the metrics (\cref{sec:metrics}).

\subsection{Tasks and runs}
\label{sec:runs}
A task instance is the tuple
\begin{equation}
\mathcal T=(w,\,g,\,s_0,\,E_0,\,A,\,B,\,O),
\label{eq:task-tuple}
\end{equation}
where $w$ is the lane's pinned source snapshots, $g$ the user goal, $s_0$ the initial state, $E_0$ the public evidence, $A$ the set of permitted tools, $B$ the probe budget, and $O$ the private oracle that the native verifier $V$ reads. Everything except $O$ is the public task of \cref{fig:framework}. A \emph{run} executes one model $m$ under one condition $c$ on one task, under one of two protocols. Protocol P0 is the released benchmark: the model receives the public task and returns one answer that the lane's native verifier checks, with no research phase and no tool or probe execution. Protocol P1, used in the representation study, has a research phase followed by an execution phase. Direct is P0 and produces the full-bank results of \cref{sec:main-results}; it is also the baseline of the representation study, whose other conditions run P1.

\subsection{Compared conditions}
We log every agent-built ontology. In a sample of 200 runs, the median ontology has 3 concepts, 11 relations and properties, and 3 Shapes Constraint Language (SHACL) node shapes and takes about 1.0k tokens, and the largest has 14 concepts and 24 relations and properties (see the supplementary material).

\label{sec:conditions}
In the research phase of P1 the agent issues probes $q_1,\dots,q_T$, each returning the label $y_t$ that the native executor would produce at a declared cost $\kappa(q_t)$, and ends with a knowledge artifact $K$ (step~4 in \cref{fig:framework})
\begin{equation}
K=\pi^{\mathrm{res}}_{m,c}\bigl(E_0,\,(q_t,y_t)_{t=1}^{T}\bigr),\qquad \textstyle\sum_{t=1}^{T}\kappa(q_t)\le B,
\label{eq:research-execution}
\end{equation}
where $\pi^{\mathrm{res}}_{m,c}$ is the research behavior of model $m$ under condition $c$. The conditions differ only in $K$, which is retrieved passages for RAG only, a lane-native declarative artifact for Native DSL (DSL in the figure), an ontology $\mathcal O$ for Ontology only, and an ontology populated from retrieved evidence for RAG + Ontology. This representation ablation keeps every artifact within a declarative subset and freezes the executor and harness (see the supplementary material). \textbf{Ontology.} An ontology $\mathcal O=(C,R,X,\Phi)$ consists of concepts $C$, typed relations $R$, constraints $X$, and cross-source mappings $\Phi$, together with provenance and declared scope; in \cref{fig:framework} it states that a Namespace requires an owner whose value matches a pattern. In the ontology conditions the agent builds its own $\mathcal O$ for each task and freezes it before execution, and a source-specific consumer interprets it when the agent acts. Artifacts in the Resource Description Framework (RDF)~\cite{rdf} and the Web Ontology Language (OWL)~\cite{owl2} use the Shapes Constraint Language (SHACL) and an OWL~RL reasoner~\cite{shacl,owlrl}, while other lanes define their own semantic operators. In Rhea--ChEBI, for example, $\Phi$ maps Rhea compounds to ChEBI identifiers that the consumer uses to answer queries.

\textbf{Baselines.} \emph{Direct} answers the task directly, with no research phase and no stored artifact. \emph{RAG only} retrieves from a fixed index over the same public evidence with pinned chunking, retriever, reranker, and top-$k$. \emph{Native DSL} writes the knowledge in the lane's own declarative language: Rego or CEL policies, DMN tables, CQL, the Systems Biology Markup Language (SBML)~\cite{sbml}, BDDL, OpenAPI schemas~\cite{openapi}, SPARQL queries~\cite{sparql} for the RDF lanes, or the closed query DSL of the RDKit lane. Unlike an ontology, which organizes typed identities, relations, constraints, and provenance behind a stable interface, a Native DSL artifact carries none of these.
\textbf{Ontology conditions.} \emph{Ontology only} builds $\mathcal O$ from the public evidence and probes. The key condition, \emph{RAG + Ontology}, retrieves evidence as RAG only does and organizes it into $\mathcal O$, so that retrieved facts are linked and constrained within the task; it tests whether a structured semantic layer adds value on top of retrieval. Index construction and ontology building are both charged to cost.
\textbf{References.} The \emph{expert-derived reference} installs the curated rule layer from which each lane's oracle is derived (ontology and shapes for the RDF lanes, the source's own rule set elsewhere) in place of the agent's artifact; it measures the pipeline with correct knowledge and is a ceiling, not a competitor. The \emph{namespace-only stub} keeps the same consumer but installs an empty ontology that declares only namespaces, which isolates the ontology's content from its interface. The stub is the initial submission of Organization and Construction tasks and is not run on Repair, where every condition starts from the defective artifact.
\textbf{Interventions.} \emph{Gold}, \emph{Corrupt}, and \emph{Repaired} pre-install the expert-derived reference artifact, a copy with one certified defect, or that copy with the defect fixed. A task is \emph{governed} if its solution depends on the corrupted element and \emph{unaffected} otherwise.

\subsection{Execution and verification}
\label{sec:success}
In the execution phase (step~5) the agent acts with the same tools in every condition under a fixed step limit and may consult $K$, which yields a trajectory $\tau$ and a final state or answer $z$. The native verifier (step~6) returns PASS if $z$ satisfies $g$ in the native executor; in the running example, OPA evaluates the Gatekeeper rule on the repaired namespaces. Each run receives two binary scores,
\begin{equation}
S=\mathbb 1\bigl[V_{\mathcal T}(z;O_{\mathcal T})=\mathrm{PASS}\bigr],\qquad S^{\mathrm v}=S\cdot\mathbb 1\bigl[\nu(\tau)=0\bigr],
\label{eq:terminal-success}
\end{equation}
where $\mathbb 1[\cdot]$ is the indicator function and $\nu(\tau)$ counts \emph{committed} policy violations, state-changing actions that the native policy forbids but that took effect; a request the policy rejects is logged but is not committed. $S^{\mathrm v}$ can differ from $S$ only on the admission-policy lanes (Gatekeeper and CEL) when state-changing actions are executed.

\subsection{Metrics}
\label{sec:metrics}
The Metrics box of \cref{fig:framework} lists four measures: task success, valid task success, mapping F1 on the Common-7 families, and cost per valid success. For model $m$ and condition $c$, let $\mathcal R_{m,c}$ be the set of all launched runs, including failed builds, exhausted budgets, and tool errors, and $N_{m,c}=|\mathcal R_{m,c}|$. The task success rate (TSR) and the valid task success rate (V-TSR) are
\begin{equation}
\begin{aligned}
\mathrm{TSR}_{m,c}&=\frac{1}{N_{m,c}}\sum_{r\in\mathcal R_{m,c}}S_r,\\
\mathrm{V\text{-}TSR}_{m,c}&=\frac{1}{N_{m,c}}\sum_{r\in\mathcal R_{m,c}}S^{\mathrm v}_r,
\end{aligned}
\label{eq:task-success}
\end{equation}
both in percent. TSR asks whether the goal was reached; V-TSR also requires that no committed violation occurred on the way.

Pooled rates over-weight lanes with many template instances, so to compare conditions $c$ and $c'$ we also weight every lane equally:
\begin{equation}
\Delta_m(c,c')=\frac{1}{|\mathcal L|}\sum_{\ell\in\mathcal L}\bigl(\mathrm{TSR}_{m,c,\ell}-\mathrm{TSR}_{m,c',\ell}\bigr),
\label{eq:paired-effects}
\end{equation}
where $\mathcal L$ is the set of 16 lanes and $\mathrm{TSR}_{m,c,\ell}$ the TSR of model $m$ under condition $c$ on lane $\ell$, in percentage points (pp). We report this lane-balanced difference next to the task-weighted difference $\mathrm{TSR}_{m,c}-\mathrm{TSR}_{m,c'}$. The interventions are summarized by the \emph{damage} (Gold minus Corrupt) and the \emph{recovery} (Repaired minus Corrupt), task-weighted on the governed and unaffected cohorts.

\paragraph{Mapping and cost metrics.} Organization lanes that emit decodable $(\mathrm{key},\mathrm{value})$ correspondences are scored with mapping F1 against independently curated gold mappings $\Phi^*$ on the eligible tasks. Unparseable outputs are excluded from F1, so it is always reported with its \emph{scoring coverage}, the fraction of eligible tasks that produced a parseable mapping. We report the Common-7 macro-F1, the mean of seven family-level micro-F1 scores over a fixed 362 tasks per model. Efficiency is the cost per valid success (CPS), $\mathrm{CPS}_m=\sum_i\mathrm{Cost}_{i,m}\big/\sum_i S^{\mathrm v}_{i,m}$ in charged input-plus-output tokens over research, execution, and failed attempts, reported with API calls and wall-clock seconds per valid success.

\paragraph{Statistical reporting.} Intervals are lane-clustered bootstrap intervals over the 16 lanes, and governed and unaffected cohorts, budgets, and validity predicates are fixed before any run.

\section{Experimental Results}
\label{sec:experiments}

\subsection{Experimental setup}
\label{sec:setup}
We evaluate nine models---Claude Opus~5.5~\cite{model-claude-opus55}, GPT6-sol~\cite{model-gpt6-sol}, Gemini~3.5~Flash~\cite{model-gemini35-flash}, DeepSeek~v4-pro~\cite{model-deepseek-v4-pro}, DeepSeek-V4.1-Flash~\cite{model-deepseek-v41-flash}, Kimi~k3~\cite{model-kimi-k3}, Qwen~3.8-max~\cite{model-qwen38-max}, Qwen~3.7~Flash~\cite{model-qwen37-flash}, and GLM~5.3~\cite{model-glm53}---on all 19{,}947 audited tasks under Direct, the released single-turn scorer, with each lane's native verifier, giving 179{,}523 model-task observations, one run per task, under one fixed prompt and budget. Every model is called through its API at its highest reasoning-effort setting and runs in the OpenCode agent harness~\cite{opencode}; results under Codex CLI~\cite{codexcli} and Claude Code~\cite{claudecode} are in the supplementary material. The representation study runs the two-phase protocol P1 of \cref{sec:runs} on the same tasks and models with Direct as its baseline, and the intervention study uses a paired cohort of 2{,}400 tasks.

\subsection{Direct baseline results}
\label{sec:main-results}
\begin{table}[t]
\centering\footnotesize
\setlength{\tabcolsep}{2.5pt}
\renewcommand{\arraystretch}{1.05}
\caption{\textbf{Direct baseline results.} Each model answers every task with the released single-turn scorer (OpenCode, maximum reasoning effort), without retrieval or a stored artifact. TSR (\%) by regime and overall on all 19{,}947 tasks; cost per valid success in tokens, API calls, and wall-clock seconds; Common-7 macro-F1 (mean F1 over the seven mapping families) and coverage (share of the 362 eligible tasks with a parseable mapping). V-TSR equals TSR under Direct and is omitted. Bold marks the best value per column.\IllustrativeCaption}
\label{tab:main-results}
\resizebox{\textwidth}{!}{%
\begin{tabular}{@{}lcccccccccc@{}}
\toprule
 & \multicolumn{5}{c}{TSR (\%) $\uparrow$} & \multicolumn{3}{c}{Cost per valid success $\downarrow$} & \multicolumn{2}{c}{Mapping (Common-7)} \\
\cmidrule(lr){2-6}\cmidrule(lr){7-9}\cmidrule(l){10-11}
Model & Application & Repair & Organization & Construction & Overall & Tokens & API calls & Wall (s) & Macro-F1 $\uparrow$ & Coverage $\uparrow$ \\
\midrule
Claude Opus 5.5 & 38.6 & 65.1 & \textbf{36.0} & 55.8 & \textbf{47.3} & 51.8k & \textbf{2.41} & 212 & \textbf{0.461} & 88.7\% \\
GPT6-sol & \textbf{38.9} & \textbf{67.4} & 28.9 & 50.3 & 45.8 & 46.7k & 2.52 & 176 & 0.443 & 87.8\% \\
Gemini 3.5 Flash & 30.8 & 60.3 & 22.6 & \textbf{57.2} & 39.6 & 19.2k & 2.95 & \textbf{36} & 0.372 & 82.9\% \\
DeepSeek v4-pro & 26.0 & 58.2 & 18.6 & 47.7 & 35.5 & 34.9k & 2.88 & 304 & 0.324 & 78.2\% \\
DeepSeek-V4.1-Flash & 24.6 & 53.8 & 24.4 & 38.0 & 34.5 & 17.9k & 3.02 & 98 & 0.291 & 74.0\% \\
Kimi k3 & 24.3 & 54.1 & 17.9 & 32.2 & 32.3 & 43.2k & 3.10 & 464 & 0.267 & 53.6\% \\
Qwen 3.8-max & 23.9 & 53.6 & 17.4 & 35.2 & 32.1 & 28.4k & 3.13 & 45 & 0.243 & \textbf{98.9\%} \\
GLM 5.3 & 22.6 & 47.5 & 16.8 & 34.8 & 29.6 & 77.9k & 3.40 & 627 & 0.286 & 28.7\% \\
Qwen 3.7 Flash & 24.5 & 44.0 & 9.6 & 36.5 & 27.4 & \textbf{16.4k} & 3.30 & 52 & 0.208 & 95.0\% \\
\bottomrule
\end{tabular}%
}

\end{table}

\begin{figure}[t]
\centering
\includegraphics[width=0.97\linewidth]{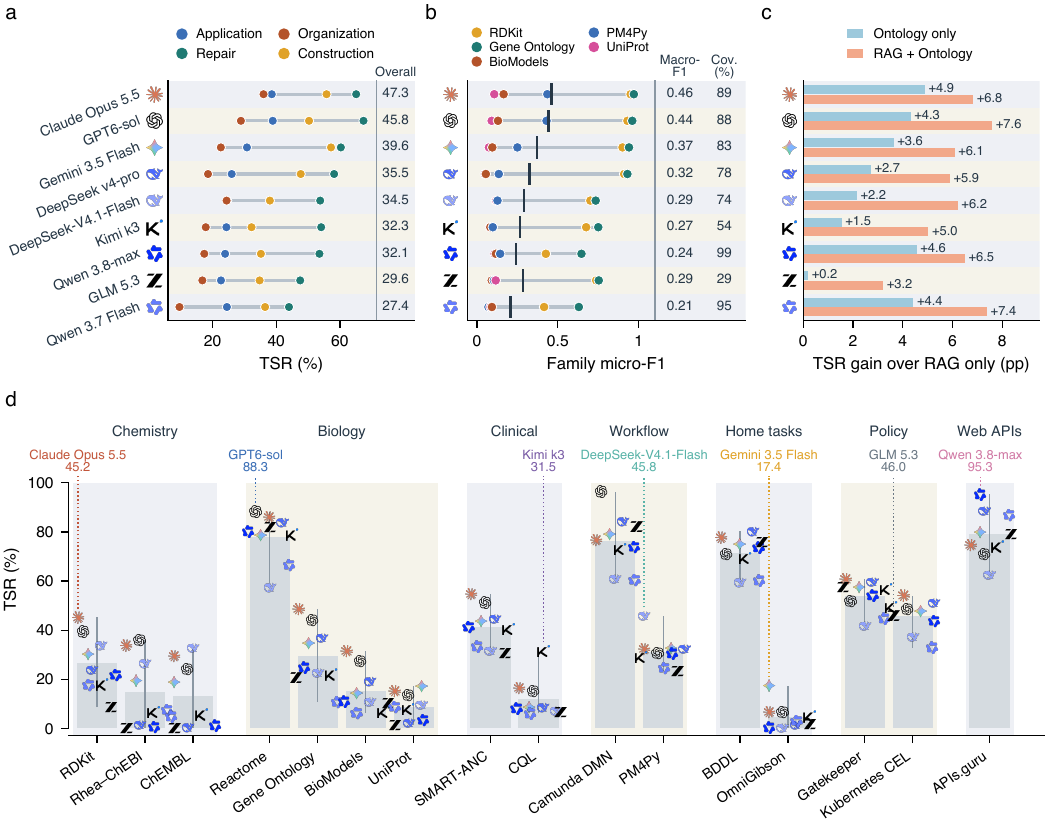}
\caption{\textbf{Per-model profiles.} Rows are models ordered by overall Direct TSR. (a,\,b)~The Direct baseline of \cref{tab:main-results}: (a)~TSR per task regime and overall; (b)~micro-F1 on the seven Common-7 mapping families, one dot per family colored by its lane, with the tick marking macro-F1 and the right columns giving macro-F1 and coverage (Cov.). (c)~The representation comparison of \cref{tab:representations}: TSR gain of Ontology only and RAG + Ontology over RAG only, in percentage points (pp). (d)~Direct TSR on the 16 lanes grouped by domain; bars give the nine-model mean and whiskers the range, and dotted callouts mark six lane leaders and GLM~5.3 on Kubernetes CEL.\IllustrativeCaption}
\label{fig:results}
\end{figure}
\Cref{tab:main-results} reports the Direct baseline (protocol P0 of \cref{sec:runs}), which measures what a model achieves with no external knowledge layer. Overall success (TSR, \cref{eq:task-success}) spans 27.4 to 47.3. Claude Opus~5.5 leads (47.3) and GPT6-sol is close (45.8), but no model is best everywhere: Claude Opus~5.5 is best on six of the 16 lanes, GPT6-sol on three, and the remaining seven are led by five different models, including two Flash models, DeepSeek~v4-pro on BDDL, and Qwen~3.8-max on APIs.guru (\cref{fig:results}d). Gemini~3.5~Flash is third overall (39.6) at less than half the token cost of the two leaders.

\subsection{Results by task regime}
\label{sec:regime-results}
Under Direct, every model succeeds most often on Repair and least often on Organization (\cref{fig:results}a). Averaged over models, Repair reaches 56.0\%, followed by Construction (43.1\%) and Application (28.3\%), while Organization reaches only 21.3\%. GPT6-sol leads Application (38.9) and Repair (67.4), Claude Opus~5.5 leads Organization (36.0), and Gemini~3.5~Flash leads Construction (57.2). Construction separates the models most (32.2 to 57.2).

\subsection{Knowledge-mapping quality}
\label{sec:mapping-results}
Mapping quality is scored on Common-7, seven mapping families from five lanes (RDKit, Gene Ontology, BioModels, PM4Py, UniProt) with 362 tasks, where a model writes cross-source correspondences that are compared with curated gold mappings. Under Direct, quality is highly uneven across families (\cref{fig:results}b). Every model reaches an F1 of at least 0.63 on Gene Ontology aspect annotation, and all but the two Qwen models reach 0.67 on RDKit molecule sets, whereas no model exceeds 0.12 on UniProt Enzyme Commission (EC) to Rhea catalyst mapping or 0.21 on the two BioModels families. Differences in macro-F1 therefore come mainly from the two PM4Py relabeling families, where Claude Opus~5.5, GPT6-sol, and Gemini~3.5~Flash reach 0.19--0.49 and no other model exceeds 0.16. GLM~5.3 reaches a macro-F1 of 0.29 but returns a scorable mapping for only 29\% of eligible tasks, whereas Qwen~3.8-max parses 99\% with a macro-F1 of 0.24, so the two must be read together.

\subsection{Knowledge representations}
\label{sec:representation-results}
\begin{table}[t]
\centering\footnotesize
\setlength{\tabcolsep}{1.7pt}
\caption{\textbf{Knowledge representations by task regime and domain.} TSR (\%) averaged over the nine models (Applic., Organ., Constr.: Application, Organization, Construction; Chem., Bio., Clin., Workf.: Chemistry, Biology, Clinical, Workflow); overall TSR is task-weighted and $\Delta$RAG is the difference to RAG only in points. Cost per valid success divides total tokens (k: thousand) or API calls, including index and ontology construction, by total valid successes. Direct reproduces \cref{tab:main-results}; bold marks the best autonomous condition per column; the expert-derived reference is a ceiling. $^\dagger$The namespace-only stub is not run on Repair, so its domain values, overall TSR, and $\Delta$RAG use the 13{,}847 other tasks.\IllustrativeCaption}
\label{tab:representations}
\resizebox{\textwidth}{!}{%
\begin{tabular}{@{}lrrrrrrrrrrrrrrr@{}}
\toprule
 & \multicolumn{4}{c}{Task regime} & \multicolumn{7}{c}{Domain} & \multicolumn{2}{c}{Overall} & \multicolumn{2}{c}{Per valid success} \\
\cmidrule(lr){2-5}\cmidrule(lr){6-12}\cmidrule(lr){13-14}\cmidrule(l){15-16}
Condition & Applic. & Repair & Organ. & Constr. & Chem. & Bio. & Clin. & Workf. & Home & Policy & Web & TSR & $\Delta$RAG & Tokens & API \\
\midrule
Direct & 28.3 & 56.0 & 21.3 & 43.1 & 16.8 & 34.8 & 30.0 & 52.3 & 57.4 & 50.8 & 79.1 & 36.0 & $-$2.7 & 37.9k & 2.91 \\
RAG only & 30.9 & 58.6 & 24.1 & 46.5 & 18.5 & 38.0 & 33.9 & 54.6 & \textbf{61.6} & 54.2 & 81.7 & 38.7 & -- & 44.3k & 2.97 \\
Native DSL & 31.6 & 59.6 & 22.7 & 47.2 & 19.9 & 37.0 & 33.4 & 55.3 & 60.3 & \textbf{55.4} & \textbf{82.4} & 39.0 & +0.3 & 51.0k & 3.46 \\
Ontology only & 30.8 & 60.6 & 33.2 & 48.8 & 26.0 & 41.6 & 32.7 & \textbf{57.9} & 57.8 & 52.1 & 78.8 & 41.9 & +3.2 & 52.0k & 3.39 \\
RAG + Ontology & \textbf{33.8} & \textbf{63.6} & \textbf{35.7} & \textbf{52.2} & \textbf{30.4} & \textbf{45.1} & \textbf{35.8} & 56.5 & 60.5 & 53.7 & 80.8 & \textbf{44.8} & +6.1 & 53.7k & 3.27 \\
\midrule
Expert-derived reference & 46.3 & 76.4 & 48.3 & 70.0 & 60.0 & 50.4 & 41.3 & 62.7 & 64.7 & 59.9 & 84.7 & 57.8 & +19.1 & 32.4k & 2.17 \\
Namespace-only stub$^\dagger$ & 23.2 & -- & 14.8 & 32.9 & 10.0 & 20.5 & 17.1 & 30.6 & 36.3 & 30.2 & 52.1 & 21.0 & $-$8.9 & 45.8k & 3.68 \\
\bottomrule
\end{tabular}%
}

\end{table}
Averaged over the nine models, TSR rises from 36.0 (Direct) to 38.7 (RAG only), 41.9 (Ontology only), and 44.8 (RAG + Ontology) (\cref{tab:representations}). The key contrast, RAG + Ontology over RAG only, is $+6.1$ points task-weighted and $+4.1$ with every lane weighted equally (\cref{eq:paired-effects}), and both lane-clustered intervals exclude zero, whereas Ontology only over RAG only ($+3.2$ task-weighted) shrinks to $+1.4$ with an interval that includes zero once lanes are weighted equally. Native DSL (39.0) is level with RAG only, and the expert-derived reference (57.8) leaves 13.0 points of headroom above RAG + Ontology. On the 13{,}847 non-Repair tasks the namespace-only stub reaches 21.0, below Direct (27.2) and Ontology only (33.7), so the gain comes from what the ontology contains. The ontology layer raises the cost per valid success from 44.3k to 53.7k tokens, and committed policy violations keep V-TSR (\cref{eq:task-success}) within 0.4 points of TSR in every condition. In a sample of 200 runs the agent-built ontologies are small, with a median of 3 concepts, 11 relations and properties, and about 1.0k tokens (see the supplementary material).

\textbf{The gain follows mapping ability, not model size.} Every model gains from the ontology layer, from $+3.2$ (GLM~5.3) to $+7.6$ (GPT6-sol) points over RAG only (\cref{fig:results}c). The two models with the lowest Common-7 coverage under Direct, GLM~5.3 (28.7\%) and Kimi~k3 (53.6\%), gain least, and for GLM~5.3 an ontology without retrieval adds only $+0.2$. Qwen~3.7~Flash, last overall but with 95.0\% coverage, gains $+7.4$, close to the leaders. An agent benefits from an ontology to the extent that it can write the mappings the ontology stores.

\textbf{Where the ontology layer pays off.} By regime the increment of RAG + Ontology over RAG only is largest on Organization (24.1 to 35.7, $+11.6$), where the cross-source mapping the agent writes is the artifact itself, and smallest on Application (30.9 to 33.8), where the source already states the rule. By domain it is largest in Chemistry ($+11.9$) and Biology ($+7.1$), led by the Rhea--ChEBI ($+14.3$), ChEMBL ($+11.8$), and UniProt ($+10.9$) lanes, whose tasks join independently curated identifiers. It is small in Clinical and Workflow ($+1.9$ each) and slightly negative in Home tasks ($-1.1$), Policy ($-0.5$), and Web APIs ($-0.9$). Native DSL is the best autonomous condition in Policy and Web APIs, where the rule is already executable in its own language. In Chemistry RAG + Ontology reaches 30.4\% against 60.0\% for the expert-derived reference, so the open problem there is acquiring the crosswalk. Per-lane values are in the supplementary material.

\subsection{Semantic-error interventions}
\label{sec:causal-results}
A paired cohort of 2{,}400 tasks per model, drawn from Repair and Organization tasks with a curated rule layer, corrupts one semantic element and then repairs it under a frozen consumer. Over the nine models, a certified defect costs 23.7 points on the 1{,}500 governed tasks, whose solution depends on the corrupted element (Gold 79.1 to Corrupt 55.5) and repair recovers 17.9, whereas the same corruption moves the 900 unaffected tasks by 0.6 points; only the governed intervals exclude zero, so the damage is specific to the corrupted element. Damage ranges from 19.6 (Kimi~k3) to 27.8 (GPT6-sol) points, and recovery is not ordered by strength: Qwen~3.7~Flash recovers 87\% of its loss and Kimi~k3 67\%. Details are in the supplementary material.

\subsection{Lane heterogeneity and cost}
\label{sec:lane-results}
Per-lane Direct success ranges from near zero to 96 (\cref{fig:results}d). Rhea--ChEBI, ChEMBL, and UniProt stay below 37\% for every model, and Rhea--ChEBI and ChEMBL alone hold 5{,}000 of the 19{,}947 tasks, which caps every model's overall rate. Single models lead individual lanes: Kimi~k3 on CQL (31.5 against at most 16.4 for the rest), DeepSeek-V4.1-Flash on PM4Py, Gemini~3.5~Flash on the OmniGibson rollouts, and Qwen~3.8-max on APIs.guru (95.3); conversely GLM~5.3 trails on RDKit (8.7) and DeepSeek-V4.1-Flash on Reactome (57.3).

Under Direct, accuracy and cost per valid success are not aligned (\cref{tab:main-results}). The three Flash models are the cheapest per valid success (16.4k--19.2k tokens), and Gemini~3.5~Flash pairs the lowest wall-clock time (36\,s) with the third-highest TSR, whereas GLM~5.3 is both the most expensive (77.9k tokens) and the slowest (627\,s). Claude Opus~5.5 pays 2.7 times the cost of Gemini~3.5~Flash per success for 7.7 more points (see the supplementary material).

\section{Conclusion}
\label{sec:conclusion}
\bench examines whether agents can turn acquired domain knowledge into reliable semantic structures that improve downstream task performance. Across nine frontier models, our results show that agent-built ontologies complement retrieval, improving task success by 6.1 points, particularly in cross-source organization and multi-source scientific domains. However, the 13.0-point gap to the expert-derived reference reveals a persistent knowledge acquisition bottleneck, while controlled semantic corruption confirms that faithful knowledge directly determines downstream reliability. These findings suggest that the central challenge for autonomous research is not merely to generate structured knowledge, but to acquire the right knowledge and make it reliably actionable. \bench provides a task-grounded foundation for evaluating this capability.

\noindent\textbf{Limitations and future work.} Our evaluation is bounded by source-native executors: clinical rule execution does not establish treatment outcomes, and SBML results remain in silico. Regime and domain effects may reflect source composition, costs are measured in tokens, and ontologies are built per task under fixed budgets. Future work will explore persistent cross-task knowledge, broader source coverage, and agents explicitly trained for autonomous knowledge acquisition.

\clearpage
\appendix
\FloatBarrier
\setcounter{figure}{0}
\setcounter{table}{0}
\renewcommand{\thefigure}{S\arabic{figure}}
\renewcommand{\thetable}{S\arabic{table}}
\renewcommand{\theHfigure}{S\arabic{figure}}
\renewcommand{\theHtable}{S\arabic{table}}
\FloatBarrier
\section*{Appendix}

This supplement gives the notation (\cref{sec:supp-notation}), the sources and the admission audit (\cref{sec:supp-sources}), one worked example per lane (\cref{sec:supp-lanes}), the task factory (\cref{sec:supp-factory}), task difficulty (\cref{sec:supp-difficulty}), the matched conditions and interventions (\cref{sec:supp-controls}), detailed representation results (\cref{sec:supp-representation}), the measurement protocol (\cref{sec:supp-measurement}).

\section{Notation}
\label{sec:supp-notation}
\begin{table}[h]
\centering
\small
\caption{\textbf{Notation.} Symbols used in Secs.~3 and 4 of the main paper. Model, representation condition, and knowledge-engineering regime are distinct axes; the regime is a property of the task.}
\label{tab:notation}
\begin{tabularx}{\textwidth}{@{}lX@{}}
\toprule
Symbol & Definition \\
\midrule
$\mathcal T=(w,g,s_0,E_0,A,B,O)$ & Task instance: pinned source snapshots of the lane, user goal, initial state, public evidence, permitted tools, probe budget, private oracle. \\
P0, P1 & The single-answer protocol of the released benchmark (Direct) and the two-phase research and execution protocol of the representation study. \\
$m,\ c$ & Model and representation condition (Direct, RAG only, Native DSL, Ontology only, RAG + Ontology, expert-derived reference, namespace-only stub, or an intervention Gold/Corrupt/Repaired). \\
$q_t,\ y_t,\ \kappa(q_t),\ T$ & Probe issued in the research phase, the label returned by the native executor, its declared cost, and the number of probes. \\
$\pi^{\mathrm{res}}_{m,c},\ K$ & Research behavior of model $m$ under condition $c$, and the knowledge artifact it produces (retrieved passages, a lane-native declarative artifact, or an ontology $\mathcal O$; Direct has none). \\
$O_{\mathcal T},\ V_{\mathcal T}(z;O_{\mathcal T})$ & Private oracle (reference solution, rules, hidden tests) and the native verifier that checks answer or final state $z$ against it with the native executor $E_w$ on the pinned snapshots $w$. \\
$\mathrm{Admit}(\mathcal T),\ \mathcal Q$ & Admission of a candidate task: the product of pass indicators over the gates (src, det, wit, mut, goal) that apply to its lane. \\
$\mathcal O=(C,R,X,\Phi)$ & Ontology artifact: concepts, typed relations, constraints, cross-source mappings. \\
$\tau,\ z$ & Trajectory (actions and observations) and final state or answer of a run. \\
$S,\ S^{\mathrm v},\ \nu(\tau)$ & Task success (verifier returns PASS), valid task success, and the number of committed policy violations; a rejected request is not a committed violation. \\
$\mathcal R_{m,c},\ N_{m,c}$ & All launched runs of model $m$ under condition $c$, including failed builds, exhausted budgets, and tool errors, and their number. \\
$\mathrm{TSR}_{m,c},\ \mathrm{V\text{-}TSR}_{m,c}$ & Task success rate and valid task success rate over $\mathcal R_{m,c}$, in percent. \\
$\mathcal L,\ \ell,\ \mathrm{TSR}_{m,c,\ell}$ & The 16 lanes, one lane, and the TSR of model $m$ under condition $c$ on that lane. \\
$\Delta_m(c,c')$ & Lane-balanced difference between conditions $c$ and $c'$, in percentage points (pp), reported next to the task-weighted difference; damage and recovery are task-weighted Gold$-$Corrupt and Repaired$-$Corrupt. \\
$\mathrm{CPS}_m,\ \mathrm{Cost}_{i,m}$ & Cost per valid success of model $m$, and the charged input-plus-output tokens of task $i$, including research, execution, index or ontology construction, and failed attempts. \\
$\Phi^*$ & Independently curated gold mappings against which mapping F1 is computed on eligible tasks, always reported with scoring coverage. \\
\bottomrule
\end{tabularx}
\end{table}

Model, representation condition, and regime are distinct axes; every artifact version an agent consults is logged.

\section{Task Sources and Admission}
\label{sec:supp-sources}
Source qualification and task admission have separate evidence requirements. Source records retain revisions, hashes, licenses, rule spans, dependencies, input types, runtime versions, and native outputs. Task records add a goal, initial state or query, tools, budget, verified solution, terminal verifier, ancestry, and public/private interface. Files, rules, mechanisms, templates, goals, instances, paraphrases, and runs are counted separately.

\paragraph{Policy and configuration sources.}
Gatekeeper supplies native admission policies~\cite{gatekeeper}. An offline resource model supports configuration tasks: native policies check each transition and a separate verifier checks the terminal goal. CEL tasks additionally qualify parameter binding, exceptions, and admission context. Both lanes use a verified nontrivial state model for multi-step tasks.

\paragraph{Chemistry sources.}
The Rhea--ChEBI lane joins independently maintained Rhea reactions and ChEBI entities through published identifiers and cross-references~\cite{rhea,chebi}; reference queries verify reaction-participant joins and exact report answers; name similarity is never used as alignment truth. RDKit patterns define a structural-selection lane~\cite{rdkit} whose answers are written in a small closed query DSL over a licensed molecule set: each retained pattern requires native matching, a vetted molecule scope, declared outputs, and independent boundary cases. ChEMBL contributes compound, assay, and target records~\cite{chembl} whose set-valued answers are checked by exact query against the pinned release.

\paragraph{Decision and biology sources.}
The Camunda DMN lane combines 700 tasks from the Camunda example tables with 300 from the DMN Technology Compatibility Kit (TCK)~\cite{camunda,dmntck}; both preserve typed inputs, native outputs, and hit-policy semantics, the TCK cases add the conformance tests that fix the engine's reading of each table, and tables are grouped by independent source and mechanism ancestry. Gene Ontology supports query, hierarchy, or repair goals within a pinned reasoning scope~\cite{go}, distinguishing entailed, contradicted, and undetermined statements. Reactome pathway membership and UniProt functional annotations~\cite{reactome,uniprot} are answered by query over pinned releases and their cross-references.

\paragraph{Numerical, clinical, and home-task sources.}
BioModels SBML tasks fix a pinned model, solver, protocol, initial conditions, and numerical tolerance. CQL and WHO SMART-ANC carry complete dependencies, terminology bindings, native execution, and approved test-patient fixtures~\cite{cql}. BEHAVIOR-1K supplies predicates and goals~\cite{bddl}, and OmniGibson embodied evaluation adds qualified observations, native simulator transitions, and terminal BDDL verification. Symbolic BDDL and embodied OmniGibson tasks occupy separate lanes.

\paragraph{Process-mining and web-API sources.}
PM4Py with licensed event logs~\cite{pm4py} supports trace-conformance tasks over Petri-net, Business Process Model and Notation (BPMN), and DECLARE models, and APIs.guru~\cite{apisguru} supports schema-grounded request validation. Organization tasks in these lanes additionally require distinct source views, independently adjudicated correspondences, and a consuming native executor.

\paragraph{Admission counts and exclusion reasons.}
For every candidate source the audit records its native-runner attempts or the concrete blocker, and exclusions for unsupported expressions, dependencies, licensing, missing truth, absent alignments, and leakage are kept apart from agent failure. Each stage has its own denominator, because one source unit can support several goals: license and native-closure exclusions count candidate source units, expressibility exclusions count native-runnable units, and missing terminal truth, triviality, and leakage are recorded at the task stage where they are tested. Admitted instances keep their base-goal ancestry. \Cref{tab:sources} lists the qualified tasks and mechanism families of each lane, and \cref{tab:audit} the task-level admission audit.

\begin{table}[H]
\centering\footnotesize
\setlength{\tabcolsep}{5pt}
\renewcommand{\arraystretch}{1.04}
\caption{\textbf{Qualification audit.} Task-level audit of 20{,}000 generated candidates under the five admission checks of \mEqAdmit{} (source integrity, native-oracle determinism, positive and negative witnesses, mutation sensitivity, and goal correctness), retaining 19{,}947 and rejecting 53 with none pending. The independent second-engine check is not available for 737 tasks, so the bank is described as passing all applicable checks rather than re-verified in full. Counts are from the released audit report.}
\label{tab:audit}
\begin{tabularx}{\linewidth}{@{}Xr@{}}
\toprule
Audit outcome / rejection reason & Count \\
\midrule
Candidates submitted & 20,000 \\
\textbf{Qualified (released)} & \textbf{19,947} \\
Pending / missing & 0 \\
\midrule
\quad Goal mismatch & 35 \\
\quad Mutation insensitive & 7 \\
\quad No positive witness & 5 \\
\quad Duplicate & 4 \\
\quad Native engine error & 2 \\
\textbf{Total rejected} & \textbf{53} \\
\midrule
Independent second-engine check NOT\_AVAILABLE & 737 \\
\bottomrule
\end{tabularx}

\end{table}

\paragraph{Source acquisition and provenance.}
Every lane is fetched from its official external source (\cref{tab:sources}) rather than a hand-written local substitute, and the exact snapshot is recorded so that a third party retrieves the same bytes rather than a later revision that could change the ground truth. A per-lane \texttt{source\_fetch\_manifest} records the resolved version, downloaded bytes, file count, indexed entries, schema pass rate, license verdict, and number of native-runnable units, and a lockfile pins the revision behind the reported results.

Downloading a tool is not acquiring knowledge. For RDKit, PM4Py, and CQL the engine alone is not a knowledge source, so each of these lanes also pins a licensed corpus of patterns, event logs, or rule libraries, and a lane is admitted only once a compliant native source is found; otherwise the audit records the concrete blocker. APIs.guru states that some definitions come from third parties, so redistribution is checked per specification rather than assumed from the repository as a whole.

\begin{table}[H]
\centering\scriptsize
\setlength{\tabcolsep}{3pt}
\renewcommand{\arraystretch}{1.05}
\caption{\textbf{Knowledge sources, native verifiers, and bank composition.} The 16 lanes draw on 18 pinned sources: Rhea--ChEBI joins two databases and Camunda DMN combines 700 Camunda example tasks with 300 DMN TCK tasks. Each lane is acquired from its official external source, pinned to a fixed revision and license, and verified by its source's native executor; the last column shows the kind of user goal the lane yields. Fam.: mechanism families; CEL: Common Expression Language; VAP: ValidatingAdmissionPolicy; OPA: Open Policy Agent; DMN: Decision Model and Notation; TCK: Technology Compatibility Kit; SBML: Systems Biology Markup Language; SMARTS: SMILES arbitrary target specification; SMILES: Simplified Molecular-Input Line-Entry System; SDF: structure-data file; CQL: Clinical Quality Language; ELM: Expression Logical Model; FHIR: Fast Healthcare Interoperability Resources; GO: Gene Ontology; GAF: GO annotation file. Task and family counts are from the released manifest.}
\label{tab:sources}
\begin{tabularx}{\textwidth}{@{}ll>{\raggedright\arraybackslash\hsize=1.0\hsize}X>{\raggedright\arraybackslash\hsize=0.8\hsize}Xrr>{\raggedright\arraybackslash\hsize=1.1\hsize}X>{\raggedright\arraybackslash\hsize=1.1\hsize}X@{}}
\toprule
ID & Domain & Lane and official source & Native executor (verifier) & Tasks & Fam. & Native knowledge acquired & Representative task \\
L01 & Policy & Gatekeeper library~\cite{gatekeeper} & OPA/Rego admission & 1,196 & 28 & Rego, ConstraintTemplate, parameters, official tests & make an object satisfy \texttt{requiredlabels} without touching guarded siblings \\
L02 & Policy & Kubernetes CEL / VAP~\cite{k8scel} & cel-go evaluation & 700 & 20 & CEL expressions, referenced schema, admission context & validate a replica bound under a CEL admission expression \\
L03 & Workflow & Camunda examples (700) + DMN TCK (300)~\cite{camunda} & DMN engine & 1,000 & 16 & DMN XML, hit policy, typed inputs, cross-table relations & read a decision table to authorize an order \\
L04 & Chemistry & Rhea + ChEBI downloads~\cite{rhea,chebi} & curated join & 3,500 & 43 & reaction RDF, participants, cross-references, ChEBI ontology & list Rhea reactions involving a given ChEBI compound \\
L05 & Chemistry & RDKit + licensed molecule set~\cite{rdkit} & SMARTS query DSL & 1,300 & 23 & functional-group hierarchy, SMARTS, SDF/SMILES entities & select molecules matching a functional-group SMARTS \\
L06 & Biology & Gene Ontology + annotations~\cite{go} & OWL RL + Jena & 2,299 & 32 & go-basic/go-plus, GAF annotations, term hierarchy & decide an entailment over the GO hierarchy \\
L07 & Biology & BioModels SBML~\cite{biomodels,sbml} & SBML simulation & 950 & 16 & model catalog, SBML, parameters, versions, protocol & repair and re-simulate an SBML model to a tolerance \\
L08 & Clinical & CQL reference implementation~\cite{cql} & CQL engine & 500 & 18 & CQL/ELM libraries, dependencies, terminology closure & evaluate a CQL eligibility expression on a fixture \\
L09 & Clinical & WHO SMART-ANC / FHIR~\cite{smartanc,fhir} & FHIR/CQL engine & 800 & 14 & PlanDefinition, Library, ValueSet, canonical references & apply an ANC decision rule to a patient fixture \\
L10 & Home tasks & BEHAVIOR-1K BDDL~\cite{bddl} & predicate check & 875 & 24 & activities, predicates, symbolic preconditions and goals & reach a symbolic BDDL goal state \\
L11 & Home tasks & OmniGibson~\cite{behavior1k} & simulator rollout & 233 & 6 & simulator stack, scene and robot assets, task instances & complete an embodied rollout to a BDDL goal \\
L12 & Workflow & PM4Py + licensed event logs~\cite{pm4py} & conformance check & 1,144 & 24 & Petri net / BPMN / DECLARE models, logs with provenance & conformance-check a trace against a process model \\
L13 & Web APIs & APIs.guru OpenAPI directory~\cite{apisguru} & schema validator & 1,150 & 19 & OpenAPI specs, origin, version, license, relations & validate a request against an OpenAPI schema \\
L14 & Chemistry & ChEMBL~\cite{chembl} & assay query & 1,500 & 14 & compound records, bioactivity assays, target annotations & retrieve the assays for a compound against a target family \\
L15 & Biology & Reactome~\cite{reactome} & pathway query & 1,500 & 12 & curated pathways, reactions, participants, cross-references & decide membership of a protein in a curated pathway \\
L16 & Biology & UniProt~\cite{uniprot} & annotation query & 1,300 & 16 & protein sequences, functional annotations, cross-references & answer a query over annotated protein function \\
\midrule
Total & & 16 lanes, 18 sources & & 19,947 & 325 & & \\
\bottomrule
\end{tabularx}
\end{table}

\section{Worked Examples per Lane}
\label{sec:supp-lanes}
\begin{figure}[!t]
\centering
\includegraphics[width=\linewidth]{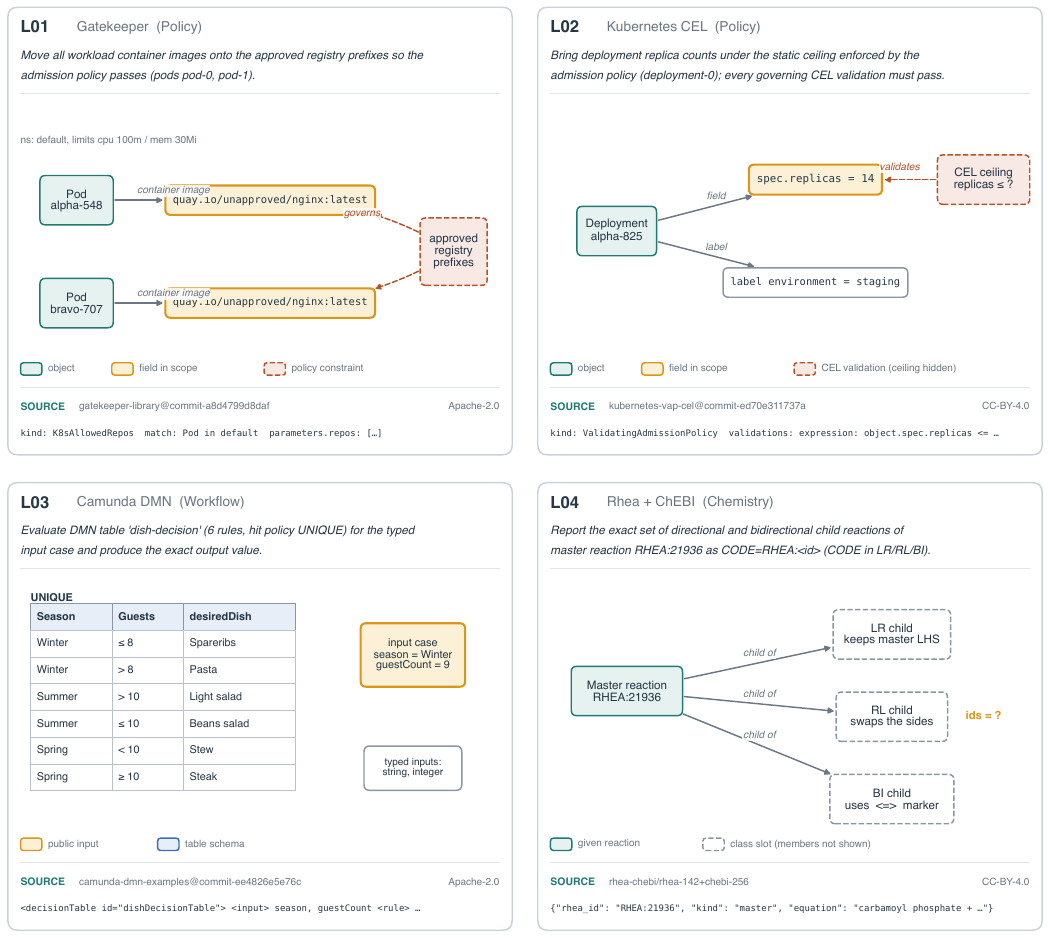}
\caption{\textbf{Lanes L01 to L04.} Gatekeeper (L01, Repair), Kubernetes CEL (L02, Repair), Camunda DMN (L03, Application), and Rhea--ChEBI (L04, Application), one public task each.}
\label{fig:lanes-a}
\end{figure}

\begin{figure}[!t]
\centering
\includegraphics[width=\linewidth]{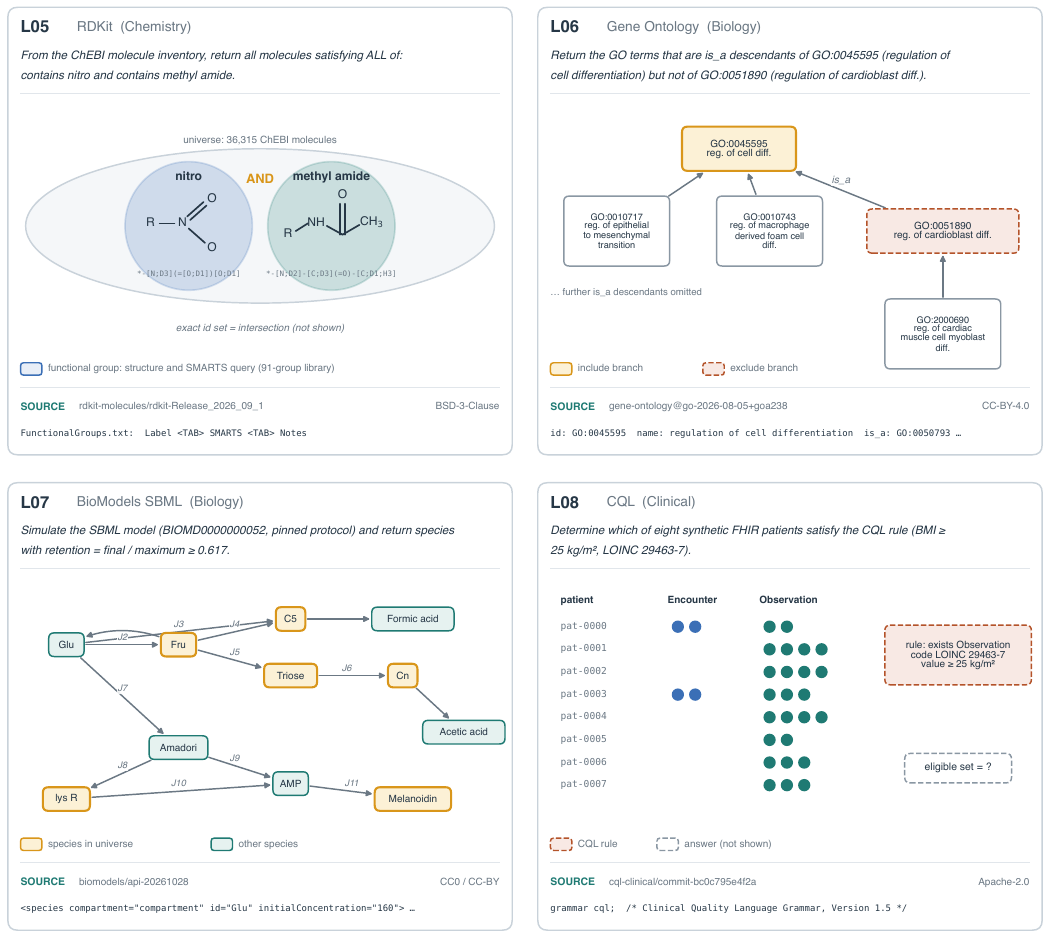}
\caption{\textbf{Lanes L05 to L08.} RDKit (L05), Gene Ontology (L06), BioModels (L07), and CQL (L08), one public task each, all in the Application regime.}
\label{fig:lanes-b}
\end{figure}

\begin{figure}[!t]
\centering
\includegraphics[width=\linewidth]{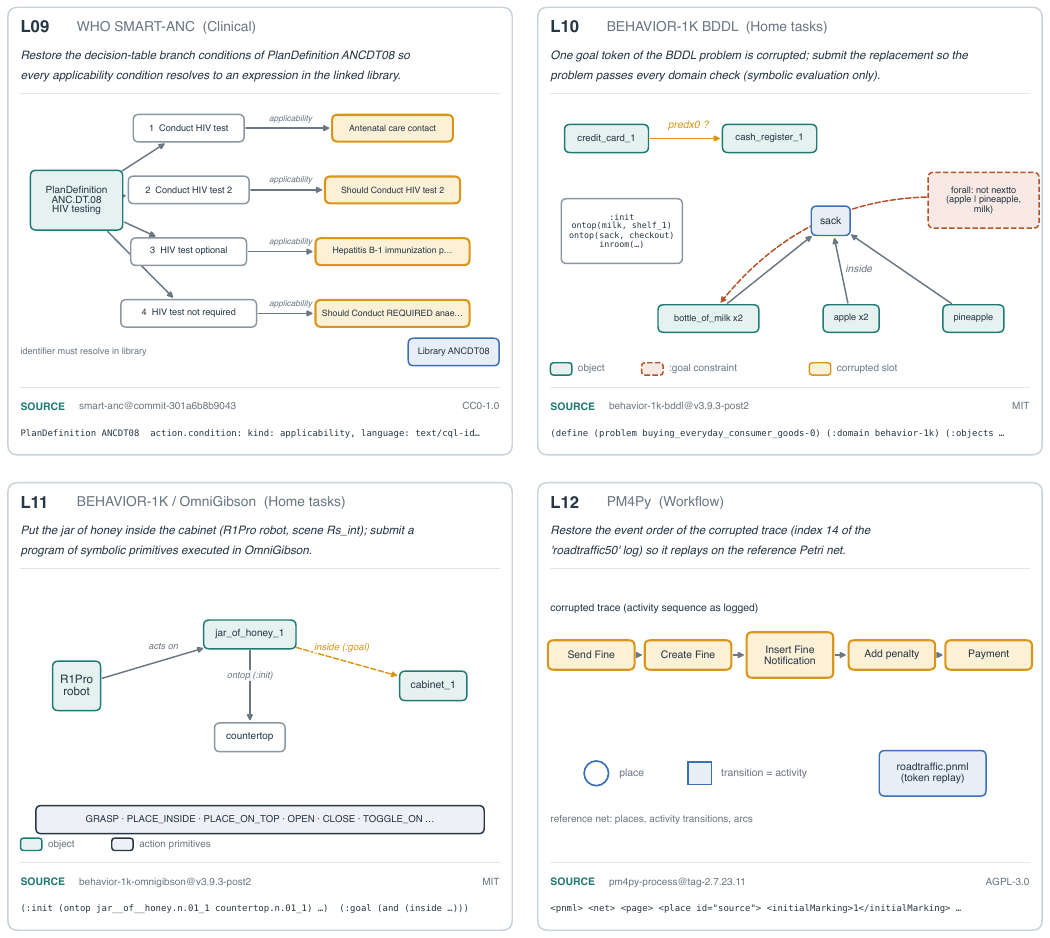}
\caption{\textbf{Lanes L09 to L12.} WHO SMART-ANC (L09, Repair), BEHAVIOR-1K BDDL (L10, Repair), OmniGibson (L11, Application), and PM4Py (L12, Repair), one public task each.}
\label{fig:lanes-c}
\end{figure}

\begin{figure}[!t]
\centering
\includegraphics[width=\linewidth]{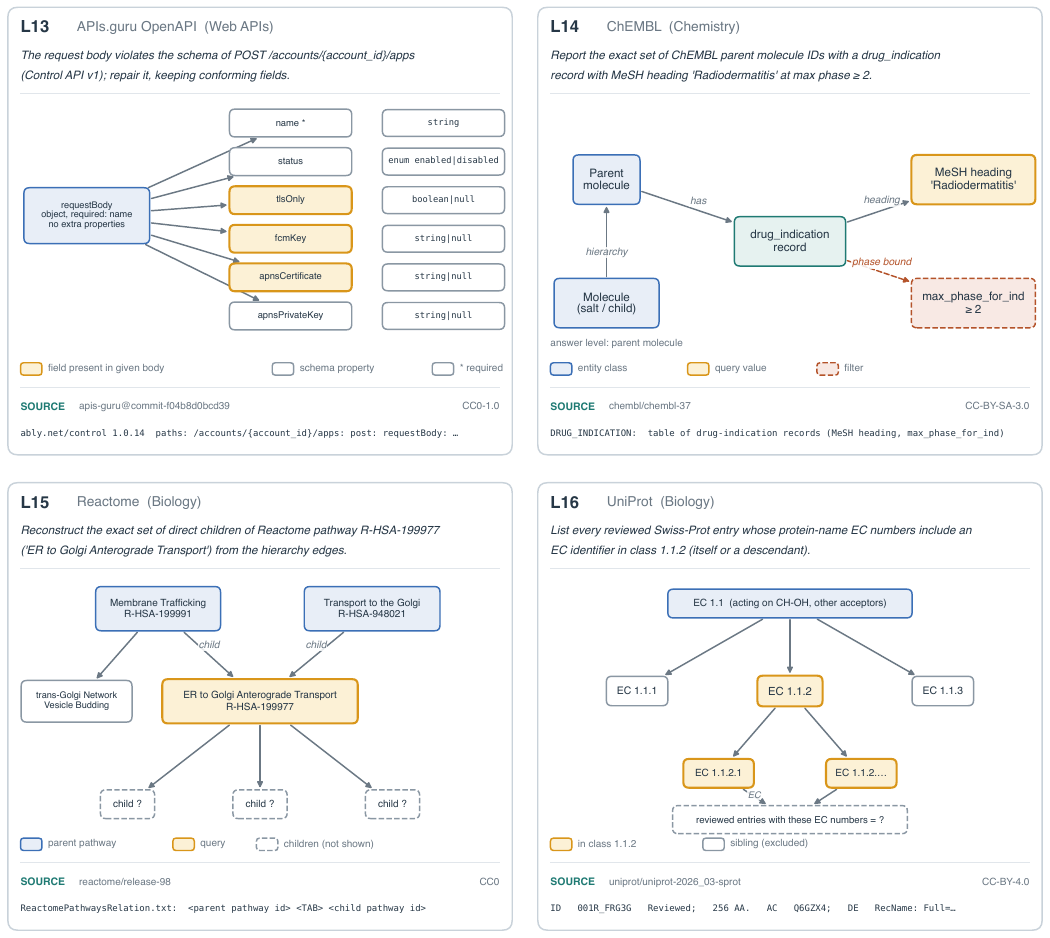}
\caption{\textbf{Lanes L13 to L16.} APIs.guru (L13, Repair), ChEMBL (L14, Application), Reactome (L15, Organization), and UniProt (L16, Application), one public task each.}
\label{fig:lanes-d}
\end{figure}

\begin{figure}[!t]
\centering
\includegraphics[width=\linewidth]{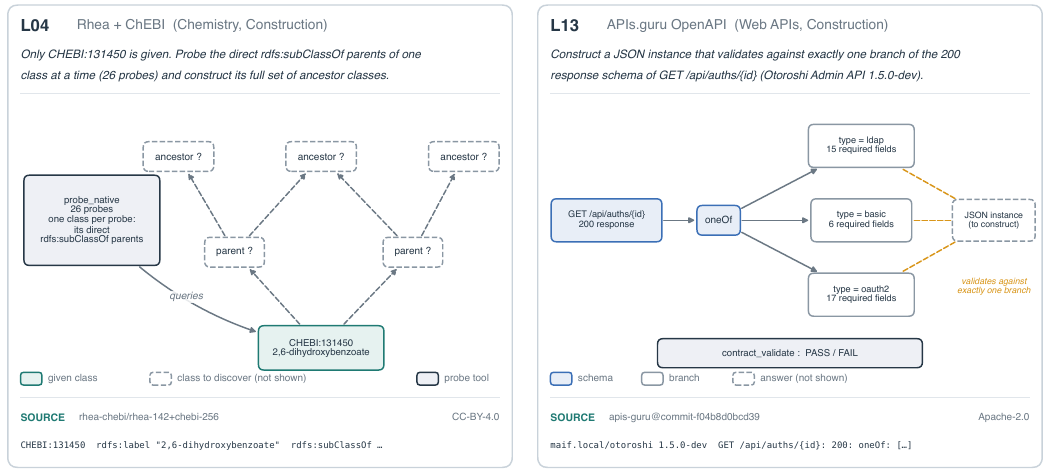}
\caption{\textbf{Two Construction tasks.} Rhea--ChEBI (L04, left): from one given ChEBI class and a budget of 26 probes that reveal direct parents, the agent constructs the full set of ancestor classes. APIs.guru (L13, right): the agent constructs a JSON instance that validates against exactly one branch of a \texttt{oneOf} response schema.}
\label{fig:construction}
\end{figure}

\begin{figure}[!t]
\centering
\begin{minipage}[t]{0.40\textwidth}\centering
\includegraphics[width=\linewidth]{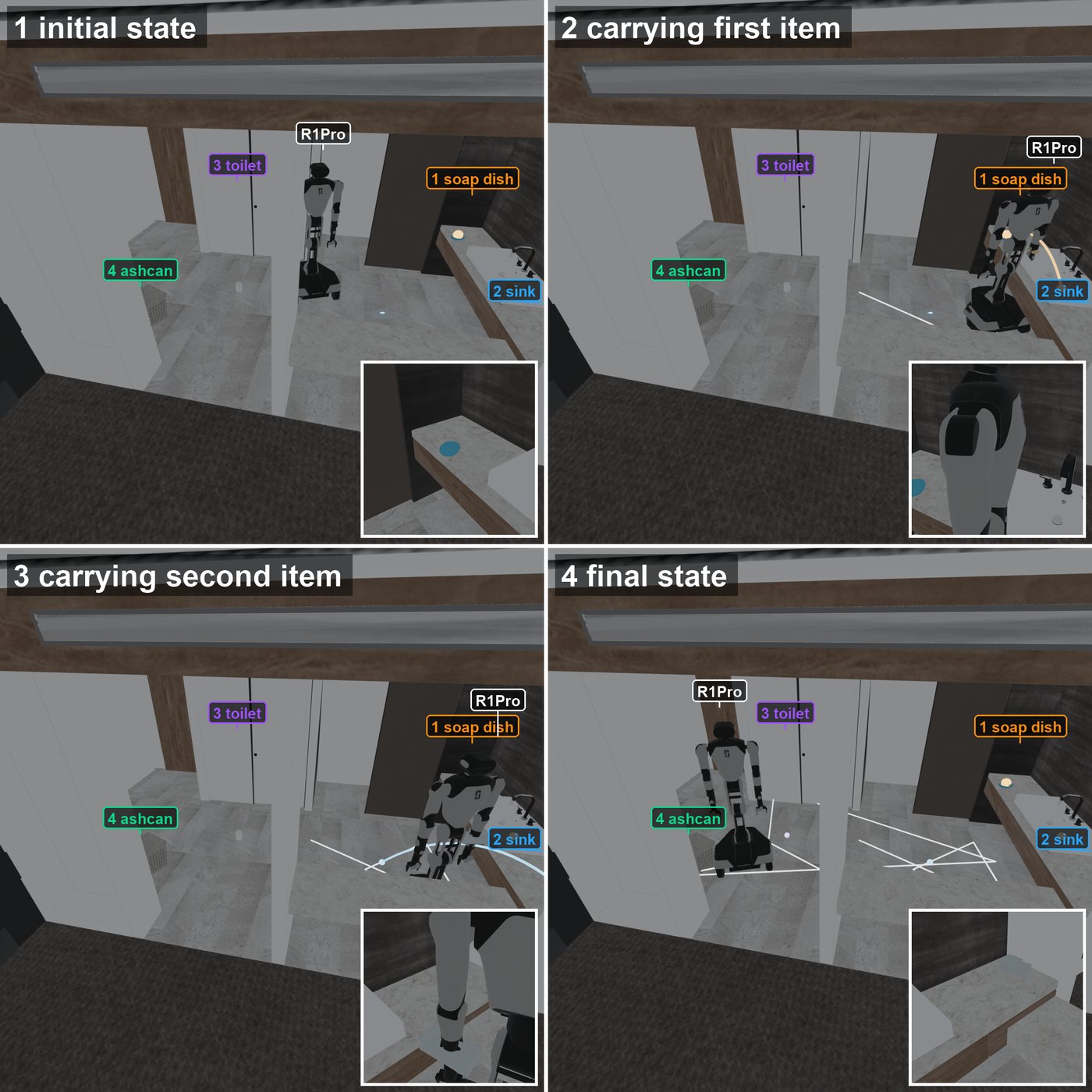}\\[1pt]{\footnotesize (a) Tidying a hotel-suite bathroom}
\end{minipage}\hfill
\begin{minipage}[t]{0.40\textwidth}\centering
\includegraphics[width=\linewidth]{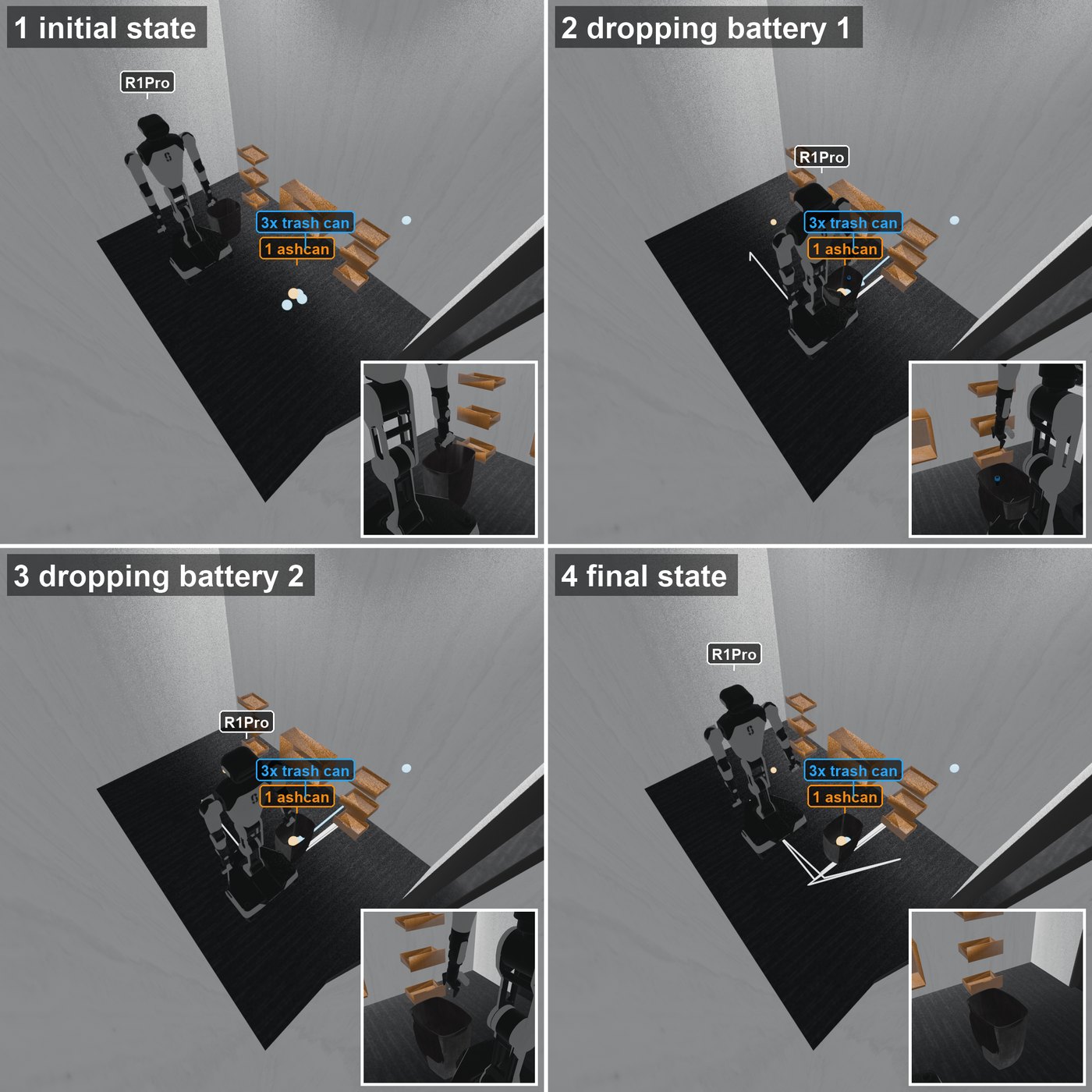}\\[1pt]{\footnotesize (b) Disposing of batteries in an office}
\end{minipage}
\caption{\textbf{Two further Application tasks of the embodied lane L11.} (a)~A robot moves four misplaced objects (bar soap, tissue dispenser, toilet tissue, and cork) to the soap dish, the sink, the toilet, and the ashcan; the reference program chains four grasp-and-place pairs. (b)~The reference program first moves the ashcan onto the floor, then places three batteries inside it. In both tasks the oracle replays the program in OmniGibson and checks the BDDL goal at the final state. Placement is replayed symbolically, so the frames are re-staged for readability and do not show contact-level physical feasibility.}
\label{fig:l11-more}
\end{figure}

\Cref{fig:lanes-a,fig:lanes-b,fig:lanes-c,fig:lanes-d} show one released task per lane, in lane order L01 to L16 (the embodied L11 task is also shown in \mFigHoney{}, and two further L11 tasks in \cref{fig:l11-more}; two Construction tasks are in \cref{fig:construction}). Each panel gives the public goal of the task and draws its public evidence as an ontology-style graph: entities and classes as boxes, relations as labeled arrows, the elements the goal asks about in amber, and policy or schema constraints dashed. Answers, verdicts, and repaired artifacts are never drawn, so a panel shows what the agent sees and must organize, not what it must produce. The captions give each task's regime: the 16 examples are nine Application, six Repair, and one Organization task, and \cref{fig:construction} adds two Construction tasks.


\section{Task Factory}
\label{sec:supp-factory}
\paragraph{Typed goals and native solutions.}
The factory constructs a source-backed task context and a typed goal before generating any language. Native solving, constrained search, or reference queries supply a solution, an existence proof, or an exact answer. Stateful instances start unsolved and replay a source-valid path; analytical instances carry set-valued, logical, or tolerance-qualified numerical answers. The terminal verifier is independent of the ontology consumer.

\paragraph{Language generation and filtering.}
An LLM writes bilingual goals and paraphrases from approved public blueprints, and any new goal structure it proposes is re-instantiated and verified natively. Deterministic filters then check source scope, schema, solvability, nontriviality, reference answers, wording consistency, duplication, and relevance.

\paragraph{Splits and contamination control.}
A \texttt{base\_goal\_id} identifies each user goal; object bindings instantiate it and paraphrases restate it. Splits group policy ancestry, cross-source homology, derived mutations, templates, and paraphrases before assignment. Each lane replaces memorable source keys with opaque identifiers and applies a counterfactual edit to the pinned source, so a memorized answer becomes wrong and success must track the edited source.

\paragraph{Public and private surfaces.}
Public tools expose approved observations, documentation, and paid probes. Private storage holds withheld rules, reference solutions, identifying labels, hidden cases, intervention eligibility, and scorer logs, behind process isolation that blocks file and path access and cleans tool responses.

\section{Task Difficulty and Stratification}
\label{sec:supp-difficulty}
For each admitted instance the generator records the constraints and cross-source links needed to verify its reference solution and, where the source supports them, a native lower bound on the action horizon and the number of source-consistent hypotheses left after public evidence. Each factor is stratified into low, medium, and high within the lanes that support it. Low Construction instances admit two hypotheses and one sufficient query, medium instances retain at least four, and high instances add interacting constraints or native prerequisites; stateful high instances require several coordinated edits while guarded sibling objects stay unchanged. The released per-task label (4{,}767 low, 8{,}446 medium, 6{,}734 high) aggregates these factors at generation time.

As a worked example, a high-difficulty Gatekeeper instance asks the agent to make five objects compliant with four pinned library rules where one pod violates three clauses at once. The native policy rejects every partial patch, so deployments, services, and pods must be fixed together while guarded sibling objects stay untouched; replica targets sit on the edges of the legal range, and illegal image names share a long prefix with the allowed registry.

\section{Matched Alternatives and Interventions}
\label{sec:supp-controls}
\paragraph{Equal information.}
RAG only and RAG + Ontology share one fixed index over the public evidence. Native DSL and the two ontology conditions author artifacts with the same primitives, interaction costs, context limits, and native verifiers, and index construction, ontology building, and use are all charged. The namespace-only stub routes decisions through an empty consumer, and the expert-derived reference replaces the agent's artifact with the curated rule layer while keeping the same planner. Paired comparisons fix model snapshots, context, tools, reset state, goals, public evidence, verifiers, and resource limits.

\begin{table}[!tb]
\centering\scriptsize
\setlength{\tabcolsep}{3pt}
\caption{\textbf{Compared conditions.} The artifact each condition stores, the state it starts from, and who consumes the artifact. The namespace-only stub is evaluated only on Organization and Construction tasks and is not run on Repair. Direct runs protocol P0; all other conditions run P1 (\mSecRuns{}).}
\label{tab:conditions}
\begin{tabularx}{\linewidth}{@{}l>{\raggedright\arraybackslash}X>{\raggedright\arraybackslash}X>{\raggedright\arraybackslash}X@{}}
\toprule
Condition & Artifact & Initial state & Consumer \\
\midrule
Direct & none & public task & model answer \\
RAG only & retrieved passages & public task & model \\
Native DSL & lane-native DSL artifact & public task & lane executor \\
Ontology only & ontology $\mathcal O$ built by the agent & public task & ontology consumer \\
RAG + Ontology & $\mathcal O$ populated from retrieved evidence & public task & ontology consumer \\
Expert-derived reference & curated rule layer & public task & frozen executor \\
Namespace-only stub & empty ontology (namespaces only) & initial submission of Organization and Construction tasks & ontology consumer \\
Gold / Corrupt / Repaired & reference artifact, one certified defect, defect fixed & pre-installed artifact & frozen executor \\
\bottomrule
\end{tabularx}
\end{table}

\paragraph{Retrieval diagnostics.} We report $\mathrm{Recall@}K(q)=|\mathrm{TopK}(q)\cap D_q^+|/|D_q^+|$ of the shared retriever on independently labelled query-passage pairs (\cref{tab:retrieval}). Recall@10 varies little across corpora (0.71--0.81), yet the gain of RAG + Ontology over RAG only ranges from $-1.1$ to $+11.9$ points across domains (\mTabRep{}) and is largest in Chemistry, where the missing piece is a cross-source crosswalk that no single passage contains.

\begin{table}[!tb]
\centering\scriptsize
\setlength{\tabcolsep}{3.5pt}
\renewcommand{\arraystretch}{1.05}
\caption{\textbf{Retrieval diagnostics by domain corpus.} Recall@5 (R@5) and Recall@10 (R@10) of the retriever shared by RAG only and RAG + Ontology on independently labelled query-passage pairs, retrieved tokens per query (Tok/q), and retrieval latency in milliseconds (ms); every corpus uses 512-token passages with a dense 768-dimensional retriever and a cross-encoder reranker.\IllustrativeCaption}
\label{tab:retrieval}
\begin{tabularx}{\linewidth}{@{}Xrrrr@{}}
\toprule
Corpus & R@5 & R@10 & Tok/q & ms \\
\midrule
Chemistry & 0.58 & 0.74 & 820 & 48 \\
Biology & 0.66 & 0.81 & 760 & 40 \\
Clinical & 0.55 & 0.71 & 840 & 51 \\
Workflow & 0.60 & 0.76 & 800 & 45 \\
Home tasks & 0.57 & 0.73 & 810 & 47 \\
Policy & 0.62 & 0.78 & 780 & 42 \\
Web APIs & 0.64 & 0.80 & 790 & 44 \\
\bottomrule
\end{tabularx}
\end{table}

\begin{figure}[!tb]
\centering
\begin{minipage}[t]{0.48\linewidth}
\centering
\includegraphics[width=\linewidth]{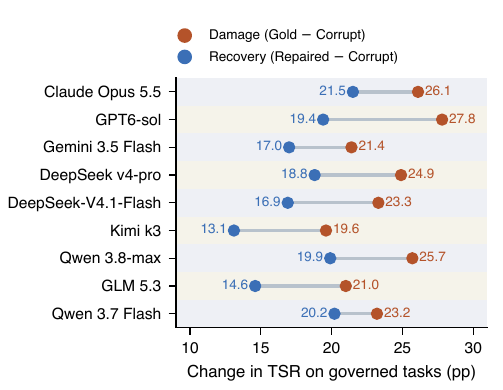}
\par\vspace{3pt}
\refstepcounter{figure}\label{fig:causal}%
{\scriptsize\textbf{(a) Semantic-error interventions per model.} Gold is the expert-derived reference artifact. Damage (Gold$-$Corrupt) and recovery (Repaired$-$Corrupt) on the 1{,}500 governed tasks, which depend on the corrupted element; pooled rates for all five conditions and both cohorts are in \cref{tab:causal}.}
\end{minipage}\hfill
\begin{minipage}[t]{0.48\linewidth}
\centering
\includegraphics[width=\linewidth]{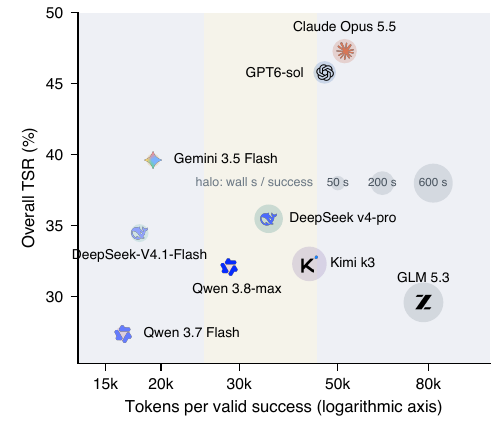}
\par\vspace{3pt}
\refstepcounter{figure}\label{fig:tsr-cps}%
{\scriptsize\textbf{(b) Accuracy against cost under Direct.} Overall TSR against tokens per valid success on a logarithmic axis; logos mark models and halo area is proportional to wall-clock seconds per success.}
\end{minipage}
\end{figure}

\paragraph{Intervention cohort.}
The intervention cohort contains 2{,}400 tasks drawn from Repair and Organization tasks whose oracle is derived from a curated rule layer, so a Gold artifact exists for each. Replay fixes actions or inputs and changes one artifact component; independent witnesses confirm the predicted semantic change, and every task is assigned to the governed (1{,}500) or unaffected (900) cohort before any run. Task-level evaluation resets the native environment and lets the planner respond to the Gold, Corrupt, Repaired, Equivalent, or Irrelevant artifact, and all assigned launches enter the effect estimates. Equivalent is a rewrite of the Gold artifact that preserves its semantics, such as renamed or reordered elements, and Irrelevant corrupts an element that the task never uses; both are controls that should leave success at the Gold level. Each model runs the same cohort, so \cref{tab:causal} pools 13{,}500 governed and 8{,}100 unaffected launches per condition over the nine models. \Cref{fig:causal} shows damage and recovery for each model.

\begin{table}[!t]
\centering
\scriptsize
\setlength{\tabcolsep}{3.5pt}
\renewcommand{\arraystretch}{1.04}
\caption{\textbf{Semantic interventions: exact counts and paired contrasts.} Success counts and rates for the 2{,}400 paired tasks per model (1{,}500 governed and 900 unaffected) under Gold (the expert-derived reference artifact), Corrupt, Repaired, Equivalent (a semantics-preserving rewrite of Gold), and Irrelevant (a defect in an element the task never uses) artifacts, pooled over the nine models. Each difference is the condition rate minus the Gold or Corrupt rate within the same subset, computed from exact counts before rounding to percentage points; Repaired$-$Corrupt is the recovery contrast.\IllustrativeCaption}
\label{tab:causal}
\begin{tabularx}{\linewidth}{@{}Xlrrrrr@{}}
\toprule
Cohort & Condition & Launches & Successes & TSR (\%) & $\Delta$ vs Gold & $\Delta$ vs Corrupt \\
\midrule
Governed & Gold & 13,500 & 10,683 & 79.13 & +0.00 & +23.67 \\
 & Corrupt & 13,500 & 7,488 & 55.47 & -23.67 & +0.00 \\
 & Repaired & 13,500 & 9,909 & 73.40 & -5.73 & +17.93 \\
 & Equivalent & 13,500 & 10,656 & 78.93 & -0.20 & +23.47 \\
 & Irrelevant & 13,500 & 10,665 & 79.00 & -0.13 & +23.53 \\
\midrule
Unaffected & Gold & 8,100 & 6,390 & 78.89 & +0.00 & +0.56 \\
 & Corrupt & 8,100 & 6,345 & 78.33 & -0.56 & +0.00 \\
 & Repaired & 8,100 & 6,354 & 78.44 & -0.44 & +0.11 \\
 & Equivalent & 8,100 & 6,390 & 78.89 & +0.00 & +0.56 \\
 & Irrelevant & 8,100 & 6,381 & 78.78 & -0.11 & +0.44 \\
\bottomrule
\end{tabularx}

\end{table}

\section{Representation Comparison and Detailed Results}
\label{sec:supp-representation}
\paragraph{Knowledge representations.} Seven conditions share tools, budget, and one frozen downstream consumer, and all but the namespace-only stub run on every task (\mSecCond{}). \mTabRep{} gives pooled, per-regime, and per-domain task success and cost; \cref{tab:rep-models} repeats \mTabMain{} for each autonomous condition and \cref{tab:domain-condition} every condition per lane. RAG + Ontology leads every autonomous condition when pooled and in Chemistry, Biology, and Clinical; Native DSL is best in Policy and Web APIs, where the rule is already executable in the lane's own language, Ontology only in Workflow, and RAG only in Home tasks. The gap to the expert-derived reference is largest in Chemistry (29.6 points), concentrated in the Rhea--ChEBI and ChEMBL crosswalks. On the 13{,}847 non-Repair tasks where it is defined, the namespace-only stub falls below Direct in every regime, so the ontology's benefit comes from its content rather than from the consumer interface. Per model, the gain of RAG + Ontology over RAG only ranges from $+3.2$ to $+7.6$ points and is smallest for the two models with the lowest Common-7 coverage.

\paragraph{Ontology statistics.} \Cref{tab:onto-stats} summarizes a sample of 200 agent-built ontologies. The median ontology has 3 concepts, 11 relations and properties, and 3 SHACL node shapes, and takes about one thousand tokens, with its shapes adding about 600; the largest has 14 concepts and 24 relations and properties.

\begin{table}[!tb]
\centering\scriptsize
\setlength{\tabcolsep}{4pt}
\caption{\textbf{Statistics of agent-built ontologies.} Median, range, and mean over a sample of 200 runs; the mean is the total divided by the number of runs (101 for mappings, which exist only on mapping tasks). Shapes are the SHACL constraints that accompany each ontology, and the bundle is the full artifact the consumer reads.}
\label{tab:onto-stats}
\begin{tabularx}{\linewidth}{@{}Xrrr@{}}
\toprule
Statistic & Median & Range & Mean \\
\midrule
Concepts & 3 & 0--14 & 5.3 \\
Object relations & 4 & 0--8 & 4.9 \\
Datatype properties & 7 & 0--17 & 10.0 \\
Relations and properties & 11 & 0--24 & 14.9 \\
SHACL node shapes & 3 & 0--12 & 4.9 \\
Property constraints & 11 & 0--25 & 12.5 \\
Mappings (101 runs) & 36 & 0--55 & 30.3 \\
Ontology size (bytes) & 3,846 & 573--23,465 & 7,163 \\
Shapes size (bytes) & 2,207 & 264--7,998 & 2,862 \\
Ontology tokens & 978 & 156--5,286 & 1,738 \\
Shapes tokens & 586 & 88--1,931 & 757 \\
Bundle tokens & 1,740 & 282--7,217 & 2,908 \\
\bottomrule
\end{tabularx}
\end{table}

\begin{table}[!t]
\centering\footnotesize
\setlength{\tabcolsep}{3pt}
\caption{\textbf{Per-model results for the four autonomous conditions.} The format of \mTabMain{} for RAG only, Native DSL, Ontology only, and RAG + Ontology: TSR (\%) by regime and overall, V-TSR, the overall gain over RAG only in points ($\Delta$RAG, the values of \mFigResults{} (panel c) for the two ontology conditions), and tokens and API calls per valid success. Each block's last row gives the mean TSR and pooled cost of \mTabRep{}. Applic./Organ./Constr.: Application, Organization, Construction.\IllustrativeCaption}
\label{tab:rep-models}
\resizebox{\textwidth}{!}{%
\begin{tabular}{@{}lrrrrrrrrr@{}}
\toprule
 & \multicolumn{6}{c}{TSR (\%)} & & \multicolumn{2}{c}{Per valid success} \\
\cmidrule(lr){2-7}\cmidrule(l){9-10}
Model & Applic. & Repair & Organ. & Constr. & Overall & V-TSR & $\Delta$RAG & Tokens & API \\
\multicolumn{10}{@{}l}{\textit{RAG only}} \\
Claude Opus 5.5 & 40.7 & 66.8 & 37.3 & 57.5 & 49.0 & 48.4 & -- & 59.7k & 2.52 \\
GPT6-sol & 40.1 & 69.8 & 31.1 & 53.6 & 47.8 & 47.5 & -- & 61.4k & 2.57 \\
Gemini 3.5 Flash & 31.9 & 62.2 & 24.9 & 59.3 & 41.3 & 41.1 & -- & 25.8k & 3.15 \\
DeepSeek v4-pro & 28.7 & 60.4 & 21.0 & 51.5 & 38.1 & 37.6 & -- & 40.7k & 3.09 \\
DeepSeek-V4.1-Flash & 29.3 & 55.8 & 26.3 & 42.1 & 37.6 & 37.3 & -- & 20.0k & 2.95 \\
Kimi k3 & 27.4 & 57.4 & 21.0 & 36.1 & 35.5 & 35.2 & -- & 47.0k & 3.13 \\
Qwen 3.8-max & 26.1 & 56.5 & 21.6 & 39.3 & 35.2 & 34.7 & -- & 32.2k & 3.13 \\
GLM 5.3 & 26.6 & 49.8 & 21.1 & 39.8 & 33.2 & 32.8 & -- & 82.6k & 3.32 \\
Qwen 3.7 Flash & 27.4 & 48.7 & 12.4 & 39.5 & 30.8 & 30.4 & -- & 20.8k & 3.15 \\
\textit{Mean / pooled} & 30.9 & 58.6 & 24.1 & 46.5 & 38.7 & 38.3 & -- & 44.3k & 2.97 \\
\midrule
\multicolumn{10}{@{}l}{\textit{Native DSL}} \\
Claude Opus 5.5 & 40.4 & 68.5 & 37.6 & 60.6 & 49.8 & 49.6 & +0.8 & 78.2k & 3.09 \\
GPT6-sol & 42.2 & 70.3 & 31.2 & 53.8 & 48.7 & 48.5 & +0.9 & 59.5k & 2.92 \\
Gemini 3.5 Flash & 33.0 & 65.6 & 24.2 & 62.1 & 42.8 & 42.6 & +1.4 & 23.9k & 3.49 \\
DeepSeek v4-pro & 29.2 & 60.2 & 19.0 & 52.0 & 37.6 & 37.5 & $-$0.4 & 54.9k & 3.60 \\
DeepSeek-V4.1-Flash & 29.0 & 55.9 & 25.9 & 42.2 & 37.4 & 37.2 & $-$0.2 & 23.2k & 3.56 \\
Kimi k3 & 27.6 & 58.5 & 19.9 & 35.7 & 35.6 & 35.4 & +0.1 & 59.4k & 3.64 \\
Qwen 3.8-max & 29.1 & 56.2 & 18.9 & 39.5 & 35.4 & 35.2 & +0.2 & 35.9k & 3.70 \\
GLM 5.3 & 25.7 & 51.6 & 18.0 & 38.6 & 32.5 & 32.4 & $-$0.7 & 93.5k & 3.94 \\
Qwen 3.7 Flash & 28.5 & 50.0 & 10.0 & 40.2 & 31.0 & 30.9 & +0.2 & 23.1k & 3.58 \\
\textit{Mean / pooled} & 31.6 & 59.6 & 22.7 & 47.2 & 39.0 & 38.8 & +0.3 & 51.0k & 3.46 \\
\midrule
\multicolumn{10}{@{}l}{\textit{Ontology only}} \\
Claude Opus 5.5 & 41.4 & 69.8 & 50.2 & 60.9 & 53.9 & 53.8 & +4.9 & 69.8k & 2.83 \\
GPT6-sol & 42.9 & 76.2 & 35.1 & 57.8 & 52.1 & 52.0 & +4.3 & 66.8k & 2.91 \\
Gemini 3.5 Flash & 34.9 & 64.2 & 31.9 & 60.6 & 45.0 & 44.9 & +3.6 & 26.9k & 3.22 \\
DeepSeek v4-pro & 26.7 & 61.2 & 33.3 & 50.3 & 40.8 & 40.5 & +2.7 & 51.0k & 3.41 \\
DeepSeek-V4.1-Flash & 26.4 & 57.8 & 35.5 & 43.5 & 39.7 & 39.6 & +2.2 & 23.1k & 3.75 \\
Kimi k3 & 23.6 & 58.1 & 31.2 & 35.4 & 37.0 & 36.8 & +1.5 & 61.4k & 3.58 \\
Qwen 3.8-max & 29.8 & 59.2 & 28.7 & 46.3 & 39.8 & 39.6 & +4.6 & 37.1k & 3.33 \\
GLM 5.3 & 22.8 & 49.0 & 27.6 & 40.9 & 33.4 & 33.3 & +0.2 & 113.0k & 4.80 \\
Qwen 3.7 Flash & 28.4 & 49.8 & 25.3 & 43.4 & 35.2 & 35.0 & +4.4 & 18.0k & 3.28 \\
\textit{Mean / pooled} & 30.8 & 60.6 & 33.2 & 48.8 & 41.9 & 41.7 & +3.2 & 52.0k & 3.39 \\
\midrule
\multicolumn{10}{@{}l}{\textit{RAG + Ontology}} \\
Claude Opus 5.5 & 41.7 & 76.1 & 49.6 & 61.8 & 55.8 & 55.7 & +6.8 & 72.4k & 2.79 \\
GPT6-sol & 43.7 & 76.9 & 45.0 & 59.8 & 55.4 & 55.2 & +7.6 & 73.6k & 2.78 \\
Gemini 3.5 Flash & 38.5 & 66.3 & 32.6 & 65.3 & 47.4 & 47.3 & +6.1 & 30.6k & 3.25 \\
DeepSeek v4-pro & 33.1 & 64.9 & 30.7 & 56.7 & 43.9 & 43.7 & +5.9 & 51.5k & 3.39 \\
DeepSeek-V4.1-Flash & 30.2 & 59.8 & 40.6 & 53.9 & 43.8 & 43.6 & +6.2 & 23.0k & 3.41 \\
Kimi k3 & 28.6 & 59.2 & 36.2 & 36.0 & 40.5 & 40.3 & +5.0 & 59.7k & 3.42 \\
Qwen 3.8-max & 29.1 & 61.5 & 35.8 & 41.3 & 41.7 & 41.6 & +6.5 & 38.7k & 3.29 \\
GLM 5.3 & 26.2 & 54.2 & 27.8 & 42.9 & 36.4 & 36.2 & +3.2 & 109.7k & 4.42 \\
Qwen 3.7 Flash & 33.5 & 53.5 & 22.9 & 52.1 & 38.2 & 37.9 & +7.4 & 20.9k & 3.11 \\
\textit{Mean / pooled} & 33.8 & 63.6 & 35.7 & 52.2 & 44.8 & 44.6 & +6.1 & 53.7k & 3.27 \\
\bottomrule
\end{tabular}%
}

\end{table}

\begin{table}[!t]
\centering\footnotesize
\setlength{\tabcolsep}{5pt}
\caption{\textbf{Representation comparison by lane.} TSR (\%) averaged over the nine models; $^\dagger$the namespace-only stub is not run on Repair, so its column is computed on the non-Repair tasks of each lane; $n$ is the number of tasks and $\Delta$ the gain of RAG + Ontology over RAG only. Direct reproduces the nine-model mean of \cref{tab:lanes-full}, and task-weighted means over the lanes of a domain reproduce the domain columns of \mTabRep{}. RAG: RAG only; DSL: Native DSL; Onto.: Ontology only; RAG+Onto.: RAG + Ontology; Ref.: expert-derived reference; Stub: namespace-only stub.\IllustrativeCaption}
\label{tab:domain-condition}
\resizebox{\textwidth}{!}{%
\begin{tabular}{@{}llrrrrrrrrr@{}}
\toprule
Domain & Lane & $n$ & Direct & RAG & DSL & Onto. & RAG+Onto. & Ref. & Stub$^\dagger$ & $\Delta$ \\
\midrule
Chemistry & Rhea--ChEBI & 3,500 & 14.7 & 16.9 & 17.4 & 27.1 & 31.1 & 64.2 & 8.6 & +14.3 \\
 & RDKit & 1,300 & 26.5 & 27.5 & 31.8 & 30.0 & 33.1 & 52.6 & 18.3 & +5.6 \\
 & ChEMBL & 1,500 & 13.1 & 14.4 & 15.1 & 19.7 & 26.1 & 56.5 & 7.8 & +11.8 \\
\addlinespace[2pt]
Biology & Gene Ontology & 2,299 & 29.5 & 32.1 & 31.4 & 34.0 & 36.4 & 40.9 & 17.8 & +4.3 \\
 & BioModels & 950 & 15.3 & 18.2 & 17.6 & 23.0 & 26.5 & 33.7 & 4.3 & +8.3 \\
 & Reactome & 1,500 & 78.0 & 81.5 & 80.3 & 84.5 & 88.9 & 91.2 & 52.9 & +7.4 \\
 & UniProt & 1,300 & 8.8 & 12.7 & 11.2 & 19.3 & 23.6 & 32.3 & 0.0 & +10.9 \\
\addlinespace[2pt]
Clinical & CQL & 500 & 12.0 & 17.2 & 16.9 & 15.3 & 19.0 & 30.6 & 2.0 & +1.7 \\
 & SMART-ANC & 800 & 41.3 & 44.4 & 43.8 & 43.6 & 46.4 & 48.0 & 26.4 & +2.0 \\
\addlinespace[2pt]
Workflow & Camunda DMN & 1,000 & 76.3 & 78.6 & 79.7 & 81.5 & 80.0 & 84.0 & 48.8 & +1.4 \\
 & PM4Py & 1,144 & 31.3 & 33.6 & 33.9 & 37.3 & 36.0 & 44.1 & 16.6 & +2.3 \\
\addlinespace[2pt]
Home tasks & BDDL & 875 & 71.4 & 75.5 & 74.0 & 71.7 & 74.3 & 77.9 & 45.5 & -1.1 \\
 & OmniGibson & 233 & 4.7 & 9.3 & 8.7 & 5.3 & 8.4 & 14.9 & 0.0 & -1.0 \\
\addlinespace[2pt]
Policy & Gatekeeper & 1,196 & 53.9 & 57.2 & 58.4 & 55.1 & 56.4 & 61.8 & 33.0 & -0.7 \\
 & Kubernetes CEL & 700 & 45.6 & 49.3 & 50.4 & 47.1 & 49.1 & 56.7 & 25.3 & -0.1 \\
\addlinespace[2pt]
Web APIs & APIs.guru & 1,150 & 79.1 & 81.7 & 82.4 & 78.8 & 80.8 & 84.7 & 52.1 & -0.9 \\
\bottomrule
\end{tabular}%
}

\end{table}

\paragraph{Per-lane detail.} \Cref{tab:lanes-full} gives task success for all nine models on all 16 lanes, grouped by domain, including Rhea--ChEBI, ChEMBL, and UniProt, where no model exceeds 36.1\%.

\begin{table}[!t]
\centering\scriptsize
\setlength{\tabcolsep}{3.5pt}
\caption{\textbf{Per-lane task success (\%) under Direct, all nine models.} TSR on each of the 16 lanes with the lane task count; bold marks the best model per lane, and the last row is the task-weighted overall of \mTabMain{}. Across lanes success spans from about 0 to 96.\IllustrativeCaption}
\label{tab:lanes-full}
\resizebox{\textwidth}{!}{%
\begin{tabular}{@{}lrrrrrrrrrr@{}}
\toprule
Source (domain) & $n$ & \rotatebox{70}{Claude Opus 5.5} & \rotatebox{70}{GPT6-sol} & \rotatebox{70}{Gemini 3.5 Flash} & \rotatebox{70}{DeepSeek v4-pro} & \rotatebox{70}{DeepSeek-V4.1-Flash} & \rotatebox{70}{Kimi k3} & \rotatebox{70}{Qwen 3.8-max} & \rotatebox{70}{GLM 5.3} & \rotatebox{70}{Qwen 3.7 Flash} \\
\midrule
Rhea--ChEBI (Chemistry) & 3,500 & 33.8 & \textbf{36.1} & 19.5 & 1.4 & 26.4 & 6.8 & 0.7 & 0.2 & 7.5 \\
RDKit (Chemistry) & 1,300 & \textbf{45.2} & 39.5 & 30.4 & 23.6 & 33.7 & 17.9 & 21.8 & 8.7 & 17.8 \\
ChEMBL (Chemistry) & 1,500 & 29.6 & 24.3 & 18.9 & 0.4 & \textbf{32.8} & 5.6 & 1.2 & 0.2 & 4.8 \\
\addlinespace[1pt]
Gene Ontology (Biology) & 2,299 & \textbf{48.6} & 44.2 & 34.7 & 36.8 & 22.6 & 22.0 & 24.8 & 20.7 & 10.7 \\
BioModels (Biology) & 950 & \textbf{31.6} & 27.4 & 14.5 & 19.1 & 10.8 & 6.6 & 11.0 & 10.2 & 6.4 \\
Reactome (Biology) & 1,500 & 86.0 & \textbf{88.3} & 78.7 & 83.8 & 57.3 & 78.9 & 80.2 & 82.0 & 66.4 \\
UniProt (Biology) & 1,300 & 15.1 & 13.6 & \textbf{17.2} & 2.2 & 9.5 & 7.8 & 3.5 & 1.4 & 8.6 \\
\addlinespace[1pt]
CQL (Clinical) & 500 & 16.4 & 15.2 & 8.6 & 8.5 & 7.0 & \textbf{31.5} & 8.1 & 6.7 & 6.1 \\
SMART-ANC (Clinical) & 800 & \textbf{54.6} & 51.3 & 43.7 & 44.6 & 31.4 & 40.4 & 41.4 & 30.7 & 33.4 \\
\addlinespace[1pt]
Camunda DMN (Workflow) & 1,000 & 76.8 & \textbf{96.4} & 79.1 & 84.2 & 60.9 & 72.9 & 73.9 & 82.2 & 60.4 \\
PM4Py (Workflow) & 1,144 & 32.4 & 30.8 & 32.8 & 32.4 & \textbf{45.8} & 29.1 & 30.5 & 23.4 & 24.5 \\
\addlinespace[1pt]
BDDL (Home tasks) & 875 & 77.7 & 71.1 & 75.2 & \textbf{80.2} & 59.4 & 69.3 & 73.6 & 75.9 & 60.2 \\
OmniGibson (Home tasks) & 233 & 6.6 & 6.6 & \textbf{17.4} & 1.3 & 0.1 & 4.8 & 0.4 & 1.7 & 3.0 \\
\addlinespace[1pt]
Gatekeeper (Policy) & 1,196 & \textbf{60.9} & 52.0 & 57.5 & 59.5 & 41.7 & 56.9 & 54.2 & 57.6 & 44.8 \\
Kubernetes CEL (Policy) & 700 & \textbf{54.1} & 48.5 & 47.8 & 51.0 & 37.3 & 49.3 & 43.7 & 46.0 & 32.9 \\
\addlinespace[1pt]
APIs.guru (Web APIs) & 1,150 & 74.6 & 70.9 & 80.1 & 88.4 & 62.5 & 74.1 & \textbf{95.3} & 79.2 & 86.9 \\
\midrule
Overall (task-weighted) & 19,947 & 47.3 & 45.8 & 39.6 & 35.5 & 34.5 & 32.3 & 32.1 & 29.6 & 27.4 \\
\bottomrule
\end{tabular}%
}

\end{table}

\paragraph{Regime by domain.} \Cref{tab:regime-domain} crosses the four regimes with the seven domains for every model under Direct. For each model the cells, weighted by their task counts, reproduce its regime TSR in \mTabMain{} and its domain TSR in \cref{tab:lanes-full}. The Organization weakness concentrates in Chemistry, where Organization TSR ranges from 1.6 (GLM~5.3) to 27.9 (Claude Opus~5.5), whereas Organization tasks in Web APIs and Home tasks are solved far more often.

\begin{table}[!t]
\centering\scriptsize
\setlength{\tabcolsep}{4pt}
\renewcommand{\arraystretch}{0.95}
\caption{\textbf{Task success by regime and domain for every model under Direct.} TSR (\%) per domain and regime; the first block (top left) gives the number of tasks per cell, and ``--'' marks cells with fewer than 30 tasks. All: the model's TSR on that domain (rows) or overall (model rows). Applic./Organ./Constr.: Application, Organization, Construction.\IllustrativeCaption}
\label{tab:regime-domain}
\begin{minipage}[t]{0.49\textwidth}
\centering
\resizebox{\textwidth}{!}{%
\begin{tabular}{@{}lrrrrr@{}}
\toprule
Model / domain & Applic. & Repair & Organ. & Constr. & All \\
\midrule
\textit{Tasks} & & & & &  \\
\quad Chemistry & 1,825 & 1,412 & 2,866 & 198 & 6,300 \\
\quad Biology & 2,168 & 1,906 & 1,635 & 340 & 6,049 \\
\quad Clinical & 559 & 528 & 105 & 108 & 1,300 \\
\quad Workflow & 786 & 742 & 394 & 222 & 2,144 \\
\quad Home tasks & 457 & 431 & 29 & 191 & 1,108 \\
\quad Policy & 778 & 734 & 28 & 356 & 1,896 \\
\quad Web APIs & 460 & 347 & 263 & 79 & 1,150 \\
\midrule
\textit{Claude Opus 5.5} & & & & & 47.3 \\
\quad Chemistry & 37.2 & 47.9 & 27.9 & 30.5 & 35.2 \\
\quad Biology & 35.5 & 66.8 & 42.7 & 48.1 & 48.0 \\
\quad Clinical & 28.7 & 54.7 & 22.4 & 42.7 & 39.9 \\
\quad Workflow & 45.9 & 67.7 & 42.6 & 48.6 & 53.1 \\
\quad Home tasks & 39.5 & 83.2 & -- & 73.0 & 62.7 \\
\quad Policy & 39.0 & 74.3 & -- & 68.3 & 58.4 \\
\quad Web APIs & 57.3 & 93.9 & 74.0 & 92.5 & 74.6 \\
\midrule
\textit{GPT6-sol} & & & & & 45.8 \\
\quad Chemistry & 33.8 & 51.8 & 25.0 & 39.2 & 34.0 \\
\quad Biology & 34.9 & 70.9 & 34.3 & 32.4 & 45.9 \\
\quad Clinical & 23.2 & 56.4 & 17.8 & 37.3 & 37.4 \\
\quad Workflow & 52.1 & 84.4 & 32.4 & 69.2 & 61.4 \\
\quad Home tasks & 48.5 & 72.4 & -- & 50.4 & 57.5 \\
\quad Policy & 38.6 & 61.9 & -- & 56.2 & 50.7 \\
\quad Web APIs & 65.6 & 98.0 & 37.5 & 93.6 & 70.9 \\
\midrule
\textit{Gemini 3.5 Flash} & & & & & 39.6 \\
\quad Chemistry & 15.7 & 41.1 & 14.6 & 38.4 & 21.6 \\
\quad Biology & 29.9 & 52.9 & 30.1 & 56.2 & 38.7 \\
\quad Clinical & 17.3 & 46.7 & 14.4 & 31.4 & 30.2 \\
\quad Workflow & 41.5 & 80.8 & 24.3 & 65.1 & 54.4 \\
\quad Home tasks & 45.0 & 80.8 & -- & 70.7 & 63.0 \\
\quad Policy & 33.7 & 75.7 & -- & 55.2 & 53.9 \\
\quad Web APIs & 73.7 & 98.0 & 62.2 & 98.0 & 80.1 \\
\midrule
\textit{DeepSeek v4-pro} & & & & & 35.5 \\
\quad Chemistry & 3.7 & 11.2 & 4.2 & 7.6 & 5.8 \\
\quad Biology & 33.2 & 52.0 & 25.6 & 53.8 & 38.2 \\
\quad Clinical & 16.3 & 50.2 & 15.0 & 26.0 & 30.7 \\
\quad Workflow & 28.4 & 98.0 & 39.7 & 47.9 & 56.6 \\
\quad Home tasks & 34.1 & 96.8 & -- & 61.7 & 63.6 \\
\quad Policy & 28.4 & 88.7 & -- & 51.9 & 56.4 \\
\quad Web APIs & 76.0 & 98.0 & 94.4 & 98.0 & 88.4 \\
\bottomrule
\end{tabular}%
}
\end{minipage}
\hfill
\begin{minipage}[t]{0.49\textwidth}
\centering
\resizebox{\textwidth}{!}{%
\begin{tabular}{@{}lrrrrr@{}}
\toprule
Model / domain & Applic. & Repair & Organ. & Constr. & All \\
\midrule
\textit{DeepSeek-V4.1-Flash} & & & & & 34.5 \\
\quad Chemistry & 20.0 & 56.4 & 22.4 & 26.1 & 29.4 \\
\quad Biology & 20.2 & 37.2 & 23.0 & 24.6 & 26.5 \\
\quad Clinical & 10.4 & 36.2 & 14.3 & 20.0 & 22.0 \\
\quad Workflow & 36.6 & 79.4 & 32.7 & 57.4 & 52.8 \\
\quad Home tasks & 28.1 & 68.3 & -- & 46.8 & 46.9 \\
\quad Policy & 33.3 & 49.3 & -- & 37.0 & 40.1 \\
\quad Web APIs & 42.1 & 98.0 & 46.5 & 78.8 & 62.5 \\
\midrule
\textit{Kimi k3} & & & & & 32.3 \\
\quad Chemistry & 8.0 & 13.0 & 7.4 & 7.5 & 8.8 \\
\quad Biology & 22.8 & 46.0 & 24.7 & 23.7 & 30.6 \\
\quad Clinical & 17.9 & 61.6 & 24.2 & 27.4 & 36.9 \\
\quad Workflow & 32.1 & 76.7 & 34.7 & 47.1 & 49.5 \\
\quad Home tasks & 40.8 & 79.5 & -- & 37.1 & 55.7 \\
\quad Policy & 28.9 & 90.5 & -- & 35.9 & 54.1 \\
\quad Web APIs & 67.1 & 98.0 & 56.5 & 67.6 & 74.1 \\
\midrule
\textit{Qwen 3.8-max} & & & & & 32.1 \\
\quad Chemistry & 3.8 & 10.8 & 3.3 & 4.9 & 5.2 \\
\quad Biology & 18.3 & 55.5 & 22.9 & 27.9 & 31.8 \\
\quad Clinical & 15.1 & 46.9 & 16.5 & 21.0 & 28.6 \\
\quad Workflow & 36.9 & 68.5 & 47.4 & 46.3 & 50.7 \\
\quad Home tasks & 35.6 & 89.4 & -- & 44.6 & 58.2 \\
\quad Policy & 29.6 & 78.5 & -- & 39.3 & 50.3 \\
\quad Web APIs & 98.0 & 98.0 & 88.8 & 89.2 & 95.3 \\
\midrule
\textit{GLM 5.3} & & & & & 29.6 \\
\quad Chemistry & 1.6 & 3.2 & 1.6 & 2.0 & 2.0 \\
\quad Biology & 19.3 & 45.8 & 27.3 & 24.9 & 30.1 \\
\quad Clinical & 14.0 & 31.0 & 16.7 & 18.4 & 21.5 \\
\quad Workflow & 39.1 & 72.9 & 36.8 & 43.6 & 50.8 \\
\quad Home tasks & 45.3 & 80.3 & -- & 52.8 & 60.3 \\
\quad Policy & 31.9 & 82.0 & -- & 41.9 & 53.3 \\
\quad Web APIs & 66.2 & 94.7 & 80.8 & 81.1 & 79.2 \\
\midrule
\textit{Qwen 3.7 Flash} & & & & & 27.4 \\
\quad Chemistry & 7.3 & 20.0 & 4.5 & 10.2 & 9.0 \\
\quad Biology & 15.5 & 44.5 & 8.4 & 27.5 & 23.4 \\
\quad Clinical & 18.0 & 29.7 & 9.7 & 27.8 & 22.9 \\
\quad Workflow & 39.3 & 55.8 & 17.9 & 40.8 & 41.2 \\
\quad Home tasks & 34.1 & 69.2 & -- & 38.7 & 48.2 \\
\quad Policy & 33.2 & 46.9 & -- & 44.8 & 40.4 \\
\quad Web APIs & 93.1 & 98.0 & 58.2 & 98.0 & 86.9 \\
\midrule
\textit{All models (mean)} & 28.3 & 56.0 & 21.3 & 43.1 & 36.0 \\
\bottomrule
\end{tabular}%
}
\end{minipage}

\end{table}

\paragraph{Valid task success.} Direct commits no state-changing actions, so V-TSR equals TSR. In the two-phase conditions an agent can commit a policy violation on the two admission-policy lanes, Gatekeeper and Kubernetes CEL (\cref{tab:vtsr}). RAG only loses the most successes to committed violations ($-4.0$ points on the two lanes), whereas Native DSL, Ontology only, and RAG + Ontology lose about 2 points because their artifacts encode the admission constraints, and the expert-derived reference loses 0.6. Over all 19{,}947 tasks the gap between TSR and V-TSR is at most 0.4 points.

\begin{table}[!t]
\centering\scriptsize
\setlength{\tabcolsep}{2.5pt}
\caption{\textbf{TSR and V-TSR on the admission-policy lanes.} Mean over the nine models for each condition on Gatekeeper and Kubernetes Common Expression Language (CEL) policies, the only lanes where a committed policy violation is possible; $\Delta$ is V-TSR minus TSR in points over both lanes. $^\dagger$Non-Repair tasks only.\IllustrativeCaption}
\label{tab:vtsr}
\begin{tabular}{@{}lrrrrrrr@{}}
\toprule
 & \multicolumn{2}{c}{Gatekeeper} & \multicolumn{2}{c}{Kubernetes CEL} & \multicolumn{3}{c}{Both lanes} \\
\cmidrule(lr){2-3}\cmidrule(lr){4-5}\cmidrule(l){6-8}
Condition & TSR & V-TSR & TSR & V-TSR & TSR & V-TSR & $\Delta$ \\
\midrule
Direct & 53.9 & 53.9 & 45.6 & 45.6 & 50.8 & 50.8 & +0.0 \\
RAG only & 57.2 & 53.2 & 49.3 & 45.1 & 54.2 & 50.2 & $-$4.0 \\
Native DSL & 58.4 & 56.3 & 50.4 & 48.7 & 55.4 & 53.5 & $-$2.0 \\
Ontology only & 55.1 & 53.6 & 47.1 & 45.0 & 52.1 & 50.4 & $-$1.8 \\
RAG + Ontology & 56.4 & 54.4 & 49.1 & 47.0 & 53.7 & 51.7 & $-$2.1 \\
\midrule
Expert-derived reference & 61.8 & 61.2 & 56.7 & 56.3 & 59.9 & 59.4 & $-$0.6 \\
Namespace-only stub$^\dagger$ & 33.0 & 30.4 & 25.3 & 21.5 & 30.2 & 27.1 & $-$3.1 \\
\bottomrule
\end{tabular}

\end{table}

\paragraph{Per-family mapping.} \Cref{tab:family-mapf1} gives the Common-7 per-family mapping F1 and scoring coverage for all nine models, the values behind \mFigResults{} (panel b).

\begin{table}[!t]
\centering\scriptsize
\setlength{\tabcolsep}{3pt}
\caption{\textbf{Common-7 per-family mapping F1 and scoring coverage under Direct.} Each cell is micro-F1 / coverage (\%), with the number of eligible tasks per family in the header (GO: Gene Ontology; BioMod.: BioModels; traj.: trajectory shape; role: species role; relab./dual: transition and dual-transition relabeling; EC: EC--Rhea catalyst mapping); macro-F1 is the mean of the seven family F1 values and coverage the share of the 362 eligible tasks with a parseable mapping. Several families stay below 0.2 F1 despite high coverage: a parseable mapping is often a wrong one.\IllustrativeCaption}
\label{tab:family-mapf1}
\resizebox{\textwidth}{!}{%
\begin{tabular}{@{}lrrrrrrrrr@{}}
\toprule
Model & \shortstack{RDKit\\sets\\(72)} & \shortstack{GO\\aspect\\(53)} & \shortstack{BioMod.\\traj.\\(40)} & \shortstack{BioMod.\\role\\(40)} & \shortstack{PM4Py\\relab.\\(35)} & \shortstack{PM4Py\\dual\\(42)} & \shortstack{UniProt\\EC\\(80)} & Macro-F1 & Cov.\ (\%) \\
\midrule
Claude Opus 5.5 & 0.950/86 & 0.970/99 & 0.210/92 & 0.120/90 & 0.470/100 & 0.400/100 & 0.107/71 & 0.461 & 88.7 \\
GPT6-sol & 0.930/83 & 0.960/98 & 0.170/90 & 0.090/88 & 0.490/100 & 0.370/100 & 0.091/73 & 0.443 & 87.8 \\
Gemini 3.5 Flash & 0.900/78 & 0.940/97 & 0.130/85 & 0.060/80 & 0.310/98 & 0.190/95 & 0.074/65 & 0.372 & 82.9 \\
DeepSeek v4-pro & 0.910/33 & 0.930/98 & 0.070/95 & 0.040/95 & 0.150/100 & 0.120/100 & 0.049/68 & 0.324 & 78.2 \\
DeepSeek-V4.1-Flash & 0.701/70 & 0.734/96 & 0.107/80 & 0.122/76 & 0.148/88 & 0.106/86 & 0.119/47 & 0.291 & 74.0 \\
Kimi k3 & 0.675/40 & 0.750/98 & 0.102/22 & 0.063/5 & 0.114/20 & 0.082/45 & 0.079/95 & 0.267 & 53.6 \\
Qwen 3.8-max & 0.426/94 & 0.646/100 & 0.124/100 & 0.110/100 & 0.159/100 & 0.128/100 & 0.109/100 & 0.243 & 98.9 \\
GLM 5.3 & 0.737/22 & 0.753/79 & 0.090/2 & 0.093/30 & 0.122/9 & 0.088/7 & 0.116/34 & 0.286 & 28.7 \\
Qwen 3.7 Flash & 0.415/92 & 0.630/100 & 0.124/96 & 0.061/94 & 0.090/98 & 0.052/98 & 0.084/92 & 0.208 & 95.0 \\
\bottomrule
\end{tabular}%
}

\end{table}

\paragraph{Accuracy and cost.} \Cref{fig:tsr-cps} plots the overall TSR and cost columns of the Direct baseline (\mTabMain{}): the three Flash models are the cheapest per valid success, and GLM~5.3 is both the most expensive and the slowest.

\paragraph{Harness sensitivity.} The main results run every model in the OpenCode harness~\cite{opencode}. \Cref{tab:harness} repeats the Direct protocol in Codex CLI~\cite{codexcli} and Claude Code~\cite{claudecode}, with each model again called through its own API at its highest reasoning-effort setting. Each vendor's model gains in its own harness: GPT6-sol rises from 45.8 to 48.0 in Codex CLI and Claude Opus~5.5 from 47.3 to 49.9 in Claude Code. The other models move in both directions, from $-3.3$ (DeepSeek-V4.1-Flash) to $+0.7$ (Qwen~3.8-max) in Codex CLI and from $-2.7$ (DeepSeek~v4-pro) to $+0.8$ (GLM~5.3) in Claude Code. Mean TSR drops only from 36.0 to 35.6 in both harnesses, but cost and latency differ: Codex CLI matches OpenCode's tokens per valid success with 9\% less wall-clock time, whereas Claude Code needs 13\% more tokens, 27\% more API calls, and 17\% more wall-clock time. GPT6-sol overtakes Claude Opus~5.5 in Codex CLI, so the leading model depends on the harness, while Qwen~3.7~Flash ranks last in all three.

\begin{table}[!t]
\centering\scriptsize
\setlength{\tabcolsep}{3pt}
\renewcommand{\arraystretch}{0.92}
\caption{\textbf{Direct results under three agent harnesses.} TSR (\%), tokens (Tok., thousands), API calls (API), and wall-clock seconds (Wall) per valid success for each model under the Direct protocol in OpenCode (default, \mTabMain{}), Codex CLI, and Claude Code; every model is called through its API at its highest reasoning-effort setting. Bold marks each model's best harness. The last row gives mean TSR and pooled cost over the nine models.\IllustrativeCaption}
\label{tab:harness}
\resizebox{\textwidth}{!}{%
\begin{tabular}{@{}lrrrrrrrrrrrr@{}}
\toprule
 & \multicolumn{4}{c}{OpenCode (default)} & \multicolumn{4}{c}{Codex CLI} & \multicolumn{4}{c}{Claude Code} \\
\cmidrule(lr){2-5}\cmidrule(lr){6-9}\cmidrule(l){10-13}
Model & TSR & Tok. & API & Wall & TSR & Tok. & API & Wall & TSR & Tok. & API & Wall \\
\midrule
Claude Opus 5.5 & 47.3 & 51.8k & 2.41 & 212 & 47.6 & 47.7k & 2.67 & 193 & \textbf{49.9} & 56.6k & 2.65 & 223 \\
GPT6-sol & 45.8 & 46.7k & 2.52 & 176 & \textbf{48.0} & 38.8k & 2.33 & 146 & 45.4 & 49.7k & 3.14 & 199 \\
Gemini 3.5 Flash & \textbf{39.6} & 19.2k & 2.95 & 36 & 39.4 & 18.8k & 3.55 & 33 & 37.7 & 25.8k & 3.86 & 44 \\
DeepSeek v4-pro & \textbf{35.5} & 34.9k & 2.88 & 304 & 34.2 & 38.9k & 3.18 & 290 & 32.8 & 37.9k & 3.91 & 423 \\
DeepSeek-V4.1-Flash & \textbf{34.5} & 17.9k & 3.02 & 98 & 31.2 & 19.9k & 4.47 & 109 & 33.8 & 21.8k & 4.52 & 102 \\
Kimi k3 & 32.3 & 43.2k & 3.10 & 464 & \textbf{32.7} & 42.3k & 3.23 & 434 & 30.5 & 50.3k & 3.72 & 533 \\
Qwen 3.8-max & 32.1 & 28.4k & 3.13 & 45 & \textbf{32.8} & 29.3k & 3.42 & 47 & 32.4 & 36.0k & 4.06 & 63 \\
GLM 5.3 & 29.6 & 77.9k & 3.40 & 627 & 28.2 & 88.9k & 4.05 & 550 & \textbf{30.4} & 80.7k & 3.88 & 734 \\
Qwen 3.7 Flash & 27.4 & 16.4k & 3.30 & 52 & 26.1 & 17.2k & 3.45 & 45 & \textbf{27.8} & 20.6k & 4.27 & 56 \\
\midrule
Mean / pooled & 36.0 & 37.9k & 2.91 & 216 & 35.6 & 37.9k & 3.28 & 198 & 35.6 & 42.9k & 3.69 & 254 \\
\bottomrule
\end{tabular}%
}

\end{table}

\section{Measurements, Sampling, and Failure Analysis}
\label{sec:supp-measurement}
\begin{table}[!t]
\centering\footnotesize
\setlength{\tabcolsep}{3pt}
\renewcommand{\arraystretch}{0.92}
\caption{\textbf{Evaluation studies.} Each study answers one question with the conditions and metrics of \mSecEval{}.}
\label{tab:studies}
\begin{tabularx}{\linewidth}{@{}l>{\raggedright\arraybackslash}X>{\raggedright\arraybackslash}X@{}}
\toprule
Study & Question & Primary measure \\
\midrule
E0 Admission & Is every released task backed by a trustworthy native oracle? & admission checks passed (\cref{tab:audit}) \\
E1 Regimes & Can agents apply, repair, organize, and construct knowledge? & TSR per regime (\mTabMain{}) \\
E2 Mapping & Are authored cross-source mappings correct where they can be scored? & Common-7 macro-F1 with scoring coverage \\
E3 Representation & Does an agent-built ontology add value on top of retrieval? & TSR; paired $\Delta$(RAG + Ontology, RAG only) (\cref{tab:ci}) \\
E4 Intervention & Does a certified semantic fault cause task loss? & damage and recovery on governed vs.\ unaffected tasks \\
E5 Lanes & Which lanes are hard for every model, and which favor one? & per-lane TSR (\cref{tab:lanes-full}) \\
E6 Efficiency & What does a valid success cost? & tokens, API calls, and wall-clock per valid success \\
\bottomrule
\end{tabularx}
\end{table}
\paragraph{Launch accounting.}
Task success requires native confirmation of the terminal goal, and valid success additionally requires no committed policy violation on the admission-policy lanes. A read-only probe belongs to the evidence interface and is never a breach. All launches stay in the denominator, including API, parsing, infrastructure, timeout, and nontermination failures.

\paragraph{Statistical units and uncertainty.}
Each task instance yields one launch per model and condition. \Cref{tab:ci} reports the headline contrasts under two estimands with lane-clustered bootstrap intervals. We cluster at the 16 lanes because lanes are the independent source compositions over which an effect must hold; clustering at the 325 mechanism families would give tighter intervals.

\begin{table}[!t]
\centering\footnotesize
\setlength{\tabcolsep}{6pt}
\renewcommand{\arraystretch}{0.92}
\caption{\textbf{Aggregate contrasts with source-clustered uncertainty.} Mean task-success difference (percentage points) between representation conditions, and between intervention conditions, over the nine models, computed two ways: \emph{task-weighted} (pooled over all 19{,}947 instances) and \emph{source-balanced} (each of the 16 lanes weighted equally, \mEqPaired{}). Brackets are 95\% lane-clustered bootstrap intervals (2{,}000 resamples); $^{*}$ marks an interval that includes zero. The first row is the key contrast of the representation study; the lower block gives the governed and unaffected intervention contrasts.\IllustrativeCaption}
\label{tab:ci}
\begin{tabular}{@{}lcc@{}}
\toprule
Contrast & Task-weighted & Source-balanced \\
\midrule
RAG + Ontology $-$ RAG only & $+6.1\,[+2.5,+9.0]$ & $+4.1\,[+1.8,+6.7]$ \\
Ontology only $-$ RAG only & $+3.2\,[+0.5,+5.8]$ & $+1.4\,[-0.4,+3.4]^{*}$ \\
Ontology only $-$ Direct & $+5.9\,[+3.4,+8.2]$ & $+4.5\,[+2.8,+6.3]$ \\
Ontology only $-$ Native DSL & $+2.9\,[+0.0,+5.3]$ & $+1.3\,[-0.7,+3.2]^{*}$ \\
Expert-derived reference $-$ RAG + Ontology & $+13.0\,[+5.5,+20.7]$ & $+9.9\,[+5.8,+15.1]$ \\
\midrule
Damage, governed (Gold$-$Corrupt) & \multicolumn{2}{c}{$+23.7\,[+21.0,+25.7]$} \\
Recovery, governed (Repaired$-$Corrupt) & \multicolumn{2}{c}{$+17.9\,[+16.1,+19.8]$} \\
Damage, unaffected (Gold$-$Corrupt) & \multicolumn{2}{c}{$+0.6\,[-0.2,+1.2]^{*}$} \\
\bottomrule
\end{tabular}

\end{table}

\paragraph{Failure taxonomy.}
Each launch receives one primary failure code under a fixed precedence rule: missing or unused artifact, knowledge mismatch between artifact and source, tool or API exception, invalid state transition, planning failure under adequate information, native-verifier fault, or residual. A terminal goal reached with a committed policy breach is a task success but not a valid success. Comparing the namespace-only stub, the agent-built conditions, and the expert-derived reference under the same consumer separates losses from missing knowledge from losses in consuming it.

\clearpage
\bibliography{main}
\bibliographystyle{iclr2027_conference}

\end{document}